%% file: main_FlexAC_camera_ready.tex
\documentclass{article}

\usepackage[final]{neurips_2025}

\usepackage[utf8]{inputenc} 
\usepackage[T1]{fontenc}    
\usepackage{url}            
\usepackage{booktabs}       
\usepackage{amsfonts}       
\usepackage{nicefrac}       
\usepackage{microtype}      
\usepackage[table,xcdraw]{xcolor}

\usepackage{enumitem}   
\usepackage{tcolorbox}      
\usepackage{graphicx} 
\usepackage{subcaption} 
\usepackage{tabularx} 
\usepackage{makecell}   
\usepackage{array}
\usepackage{multirow}
\usepackage{amsmath} 
\usepackage[pagebackref=true,breaklinks=true,colorlinks,bookmarks=true,bookmarksopen=false,bookmarksnumbered=true]{hyperref}
\usepackage{wrapfig}
\usepackage{cleveref}
\crefname{wraptable}{Table}{Tables}
\usepackage{xspace}  

\newcommand{\stoptocwriting}{%
  \addtocontents{toc}{\protect\setcounter{tocdepth}{-5}}}
\newcommand{\resumetocwriting}{%
  \addtocontents{toc}{\protect\setcounter{tocdepth}{\arabic{tocdepth}}}}

\makeatletter  
\DeclareRobustCommand\onedot{\futurelet\@let@token\@onedot}
\def\@onedot{\ifx\@let@token.\else.\null\fi\xspace}

\def\eg{\emph{e.g}\onedot}

\def\etal{\emph{et al}\onedot}
\makeatother

\title{
  \textbf{FlexAC}\raisebox{-0.18\height}{\includegraphics[height=1.1em]{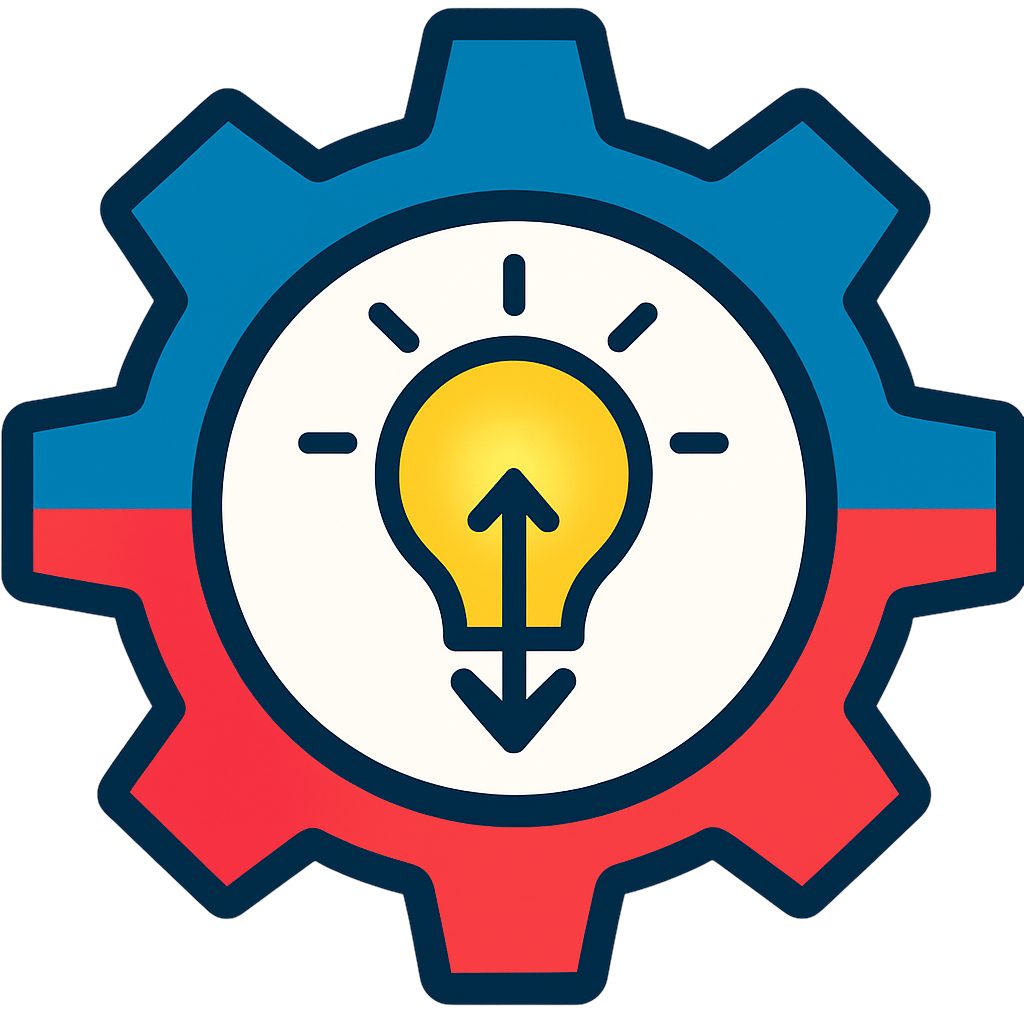}}\xspace: Towards Flexible Control of Associative Reasoning in Multimodal Large Language Models
}

\author{%
  Shengming Yuan\textsuperscript{1}\thanks{Equal contribution.} \\
  \texttt{\small shengming.yuan@outlook.com} \\
  \And 
  Xinyu Lyu\textsuperscript{2}\footnotemark[1] \\
  \texttt{\small xinyulyu68@gmail.com} \\
  \And 
  Shuailong Wang\textsuperscript{1} \\
  \texttt{\small wslliongliong@gmail.com} \\
  \AND 
  Beitao Chen\textsuperscript{1} \\
  \texttt{\small chenbeitao@gmail.com} \\
  \And 
  Jingkuan Song\textsuperscript{3} \\
  \texttt{\small jingkuan.song@gmail.com} \\
  \And 
  Lianli Gao\textsuperscript{1}\thanks{Corresponding author.} \\
  \texttt{\small lianli.gao@uestc.edu.cn} \\
  \AND 
  \vspace{-1em}
    \\ 
  \textsuperscript{1}University of Electronic Science and Technology of China \\
  \textsuperscript{2}Southwestern University of Finance and Economics, Chengdu, China \\
  \textsuperscript{3}Tongji University
}

\begin{document}

\maketitle

\stoptocwriting 

\begin{abstract}

    Multimodal large language models (MLLMs) face an inherent trade-off between \emph{faithfulness} and \emph{creativity}, as different tasks require varying degrees of \emph{associative reasoning}. 
    However, existing methods lack the flexibility to modulate this reasoning strength, limiting MLLMs' adaptability across factual and creative scenarios.
    To bridge this gap, we propose equipping MLLMs with mechanisms that enable \textbf{\textit{flexible control over associative reasoning}}.
    We begin by investigating the internal mechanisms underlying associative behavior in MLLMs and  \textbf{find that:} 
    \textbf{(1)} middle layers play a pivotal role in shaping model’s associative tendencies,  \textbf{(2)} modifying representations in these layers effectively regulates associative reasoning strength, and  \textbf{(3)} hallucinations can be exploited to derive steering vectors that guide this modulation. 
    Building on these findings, we introduce \textbf{Flex}ible \textbf{A}ssociation \textbf{C}ontrol (\textbf{FlexAC}), a \textbf{lightweight and training-free} framework for modulating associative behavior in MLLMs. 
    FlexAC first induces hallucination-guided intermediate representations to \textbf{encode associative directions}. 
    Then, it selects high-association instances to construct effective associative steering vectors, whose strengths are \textbf{adaptively calibrated} to balance creative guidance with output stability. 
    Finally, recognizing the multi-dimensional nature of associative reasoning, FlexAC incorporates task-specific associative vectors derived from a forward pass on a few target-domain samples, enabling models to follow \textbf{diverse} associative directions and better adapt to \textbf{creative tasks}. Notably, our method achieves up to a 5.8× improvement in creativity on Creation-MMBench and a 29\% reduction in hallucination rate on CHAIR, surpassing existing baselines and demonstrating its effectiveness in enabling flexible control over associative reasoning in MLLMs. Our code is available at \url{https://github.com/ylhz/FlexAC}.

\end{abstract}

\section{Introduction}
\label{sec:intro}
In cognitive science, divergent and convergent thinking represent two distinct modes of human associative behavior: convergent thinking relies on typical, fact-based associations to support faithful reasoning, whereas divergent thinking engages atypical, context-dependent associations to foster creativity~\cite{gabora2018neural}. 
Recent studies show that multimodal large language models (MLLMs)~\cite{liu2023llava,bai2023qwenvl,wu2024deepseekvl2mixtureofexpertsvisionlanguagemodels} exhibit brain-like properties, such as structured embedding spaces~\cite{goldstein2025unified}, cross-modal integration~\cite{tang2023brain}, and higher-order cognitive functions~\cite{jiang2024survey}, indicating that they emulate human associative processes.
Consequently, like the human brain, \textbf{MLLMs require the capacity to flexibly regulate associative reasoning strength to support both faithful reasoning and creative generation.}

\begin{figure}[t]
  \begin{center}
  \includegraphics[]{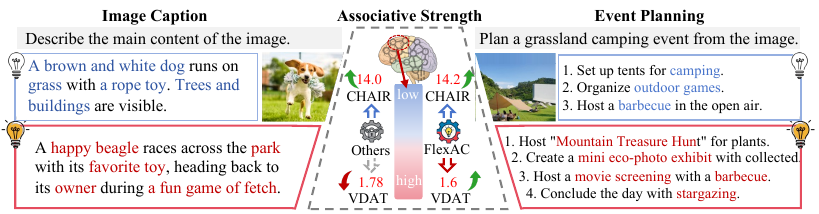}
  \end{center}
  \caption{\textbf{Different tasks require different levels of associative reasoning}: factual tasks (\eg, image caption) benefit from lower association, while creative tasks (\eg, event planning) thrive on higher association. Existing methods suppress hallucinations at the cost of creativity (\eg, -1.78 on VDAT; "Others" from Ha-DPO). FlexAC enables MLLMs to adjust associative reasoning strength accordingly.}
       \vspace{-2em}
  \label{fig:intro}
\end{figure}

However, existing methods lack the flexibility to modulate associative reasoning strength, limiting MLLMs' adaptability across factual and creative scenarios. 
On one hand, current hallucination mitigation techniques, such as Contrastive Decoding~\cite{leng2024vcd,wang2024icd,lyu2024alleviating} and Direct Preference Optimization~\cite{zhao2023hallucinations}, focus on improving faithfulness but often suppress associative reasoning capabilities, thereby hindering performance on tasks involving imaginative understanding and literary expression. 
On the other hand, how to enhance MLLMs' creativity in a controllable and task-specific manner remains underexplored.
For instance, as illustrated in \Cref{fig:intro}, existing hallucination mitigation techniques improve faithfulness (14.0 ↓ in CHAIR) but lack mechanisms for enhancing creativity, resulting in reduced associative reasoning strength (1.78 ↓ in VDAT) and poor performance on tasks such as event planning.
This gap highlights \textbf{the need for equipping MLLMs with controllable mechanisms to flexibly modulate associative reasoning strength based on task demands.}

To enable controllable modulation of associative reasoning strength, we begin by examining how associative behavior emerges within MLLMs. Drawing inspiration from prior works~\cite{rimsky2024caa, chuang2024dola}, we hypothesize that hallucination and creativity arise from shared associative mechanisms, whose manifestations vary with task demands. To validate this, we collect input-response pairs containing both grounded (low-association) and hallucinated (high-association) outputs, and analyze their internal representations to uncover how associative behavior is reflected within the model.
Our analysis reveals three key findings (see ~\Cref{sec:analyze-layer-localization} and ~\Cref{sec:analyze_control_strategy}): 
(1) Associative behaviors are primarily encoded in the middle layers, where the representations of grounded and hallucinated responses become distinctly separable; 
(2) Modifying internal representations at these layers can effectively alter the strength and direction of associative reasoning; 
(3) Direction of hallucinated representations can stimulate associative reasoning capability, offering a potential control signal for this modulation. 
\textbf{These findings indicate that associative tendencies are encoded in middle layers and can be modulated through targeted interventions guided by hallucination.}

Motivated by these findings, we propose \textbf{Flex}ible \textbf{A}ssociation \textbf{C}ontrol (\textbf{FlexAC}), a lightweight and training-free framework for modulating associative behavior in MLLMs. 
The core idea is to first extract the associative vector from hallucinated responses (Phase I: Offline Control Vector Construction), which exhibit strong associative tendencies, 
and then apply it at inference time to guide model behavior (Phase II: Inference-Time Control). 
\textbf{In the Offline Control Vector Construction Phase}, FlexAC performs \textbf{three key steps:}
(1) Hallucination-Guided Intermediate States: We collect grounded–hallucinated response pairs, and measuring the differences between their hidden states within model’s middle layers, which encode the associative direction.
(2) Instance Selection: To reduce noise from individual samples, we select the top-K response pairs with the largest association shifts and average their differences to obtain a reliable steering vector.
(3) Directional Integration: To further support tasks requiring multi-dimensional associations (e.g., storytelling or metaphor generation), we augment the general associative vector with task-specific associative vectors derived from GPT-4o-generated, high-association samples. These vectors are incorporated at inference time for fine-grained and controllable modulation. 
\textbf{In the Inference-Time Control Phase}, we apply the combined steering vector during inference. 
However, uniformly applying this vector can lead to over-steering, especially for inputs already exhibit strong associative behavior, causing irrelevant outputs or stylistic drift.
To mitigate this, we introduce \textbf{Steering Intensity Calibration}, which adaptively scales the steering vector: amplifying it when associative behavior is weak, and attenuating it when the desired level has been reached. 

To evaluate the effectiveness of FlexAC in controlling associative behavior, we conduct experiments across three fronts: 
hallucination mitigation (CHAIR~\cite{rohrbach2018CHAIR} and POPE~\cite{li2023POPE} for low-association tasks), creativity enhancement (VDAT and Creation-MMBench~\cite{fang2025creaion-mmbench} for high-association tasks), and general-purpose evaluation (MME~\cite{fu2023mme}, MMMU~\cite{yue2023mmmu}, and MMStar~\cite{chen2024mmstar}). 
Results show that FlexAC enables flexible modulation of associative reasoning capability, achieving state-of-the-art performance on both low- and high-association tasks while enhancing general capabilities. 

In summary, our contributions are fourfold: (1)We present a unified perspective that links hallucination and creativity to associative reasoning, identifying middle-layer representations as key control points. 
(2)We propose \textbf{FlexAC}, a lightweight and training-free framework for flexible modulation of associative strength, enabling task-aware switching between hallucination suppression and creativity enhancement. 
(3)We introduce VDAT, a benchmark specifically designed to evaluate associative reasoning strength.
(4)We conduct comprehensive experiments demonstrating that FlexAC effectively controls associative behavior across hallucination, creativity, and general-purpose benchmarks.

\section{Analyzing and modulating associative behavior in MLLMs}

\subsection{Analyzing layer-wise localization of associative processes}
\label{sec:analyze-layer-localization}

\paragraph{Feature Distance Analysis: Quantifying layer-wise differences between associative and non-associative representations.} 
To identify where associative behavior emerges, we analyze layer-wise representations in LLaVA-1.5-7b using 1000 images from COCO2024. 
For each image, we collect two type of responses: a grounded (non-associative) response from the model’s default output, and a hallucinated (associative) response induced via blurred inputs and specific prompts~\cite{leng2024vcd}. 
Here, we use hallucinated responses to represent associative behavior, as they often include many imaginative contents, objects that do not exist in the image but are semantically related to the scene, reflecting the model’s associative tendencies. 
We then extract the associative features $f^{(a)}$ and non-associative features $f^{(n)}$ from all intermediate layers for both data types (visualized in \Cref{fig:fea_vis_pca}). The full data construction and feature extraction process is detailed in 
Appendix ~\ref{exp_appendix:data_generate}. 
Next, we compute the cosine distance and Euclidean distance between $f^{(a)}$ and $f^{(n)}$ across all layers. 
The cosine distance $\mathcal{D}_{\text{cos}}$ is used to evaluate the directional alignment between associative and non-associative features, while Euclidean distance $\mathcal{D}_{\text{Euc}}$ measures the spatial distribution differences.

\begin{figure}[h]
    \centering
    \subfloat[Cosine distance]{
        \includegraphics[width=0.23\linewidth]{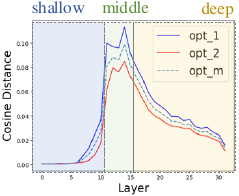}
        \label{fig:feature_diff_1}
    }
    \hfill
    \subfloat[Euclidean distance]{
        \includegraphics[width=0.23\linewidth]{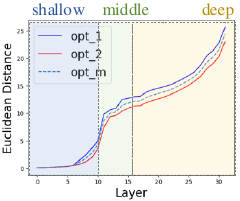}
        \label{fig:feature_diff_2}
    }
    \hfill
    \subfloat[Cosine distance]{
        \includegraphics[width=0.23\linewidth]{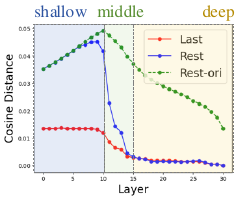}
        \label{fig:feature_diff_3}
    }
    \hfill
    \subfloat[Euclidean distance]{
        \includegraphics[width=0.23\linewidth]{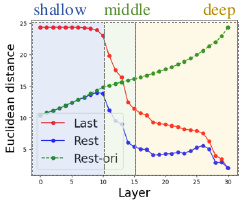}
        \label{fig:feature_diff_4}
    }
    \caption{(a) and (b) show the cosine and Euclidean distances between associative and non-associative features across layers. (c) and (d) illustrate the impact of replacing associative features in different layers on subsequent layers.``Last'' and ``Rest'' denotes the final layer difference $d_L$ and the average layer difference $\bar{d}_{m:L}$, respectively. ``Rest-ori'' represents the original mean feature distance $\bar{d}_{m:L}$ without replacement.}
    \label{fig:feature_diff}
\end{figure}

\begin{figure}[ht]
    \centering
    \begin{minipage}[t]{0.47\linewidth}
        \centering

        \includegraphics[width=0.85\textwidth]{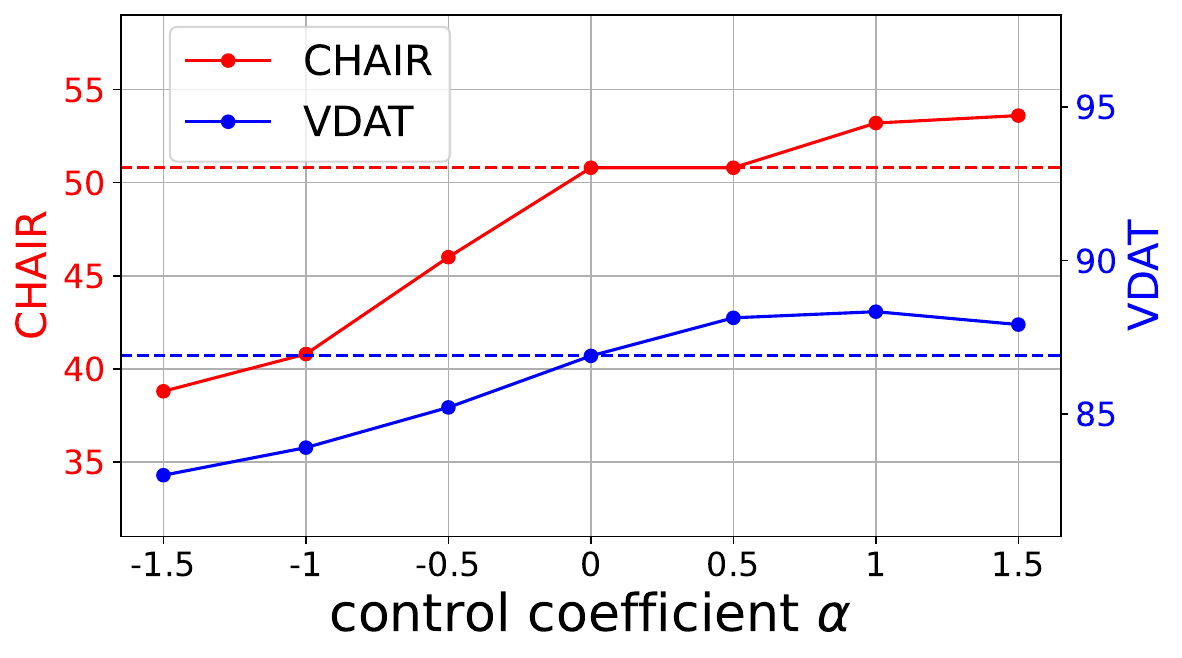}
        \caption{\textbf{Impact of Middle Layer Control on Hallucination-Driven Behavior.}
        Adjusting $\alpha$ increases both hallucination (CHAIR) and creativity (VDAT), suggesting that associative strength can be modulated through middle-layer control using hallucination representations.
        }
        \label{fig:factor}
        
    \end{minipage}
    \hfill
    \begin{minipage}[t]{0.51\linewidth}
        \centering
        \includegraphics[width=\linewidth]{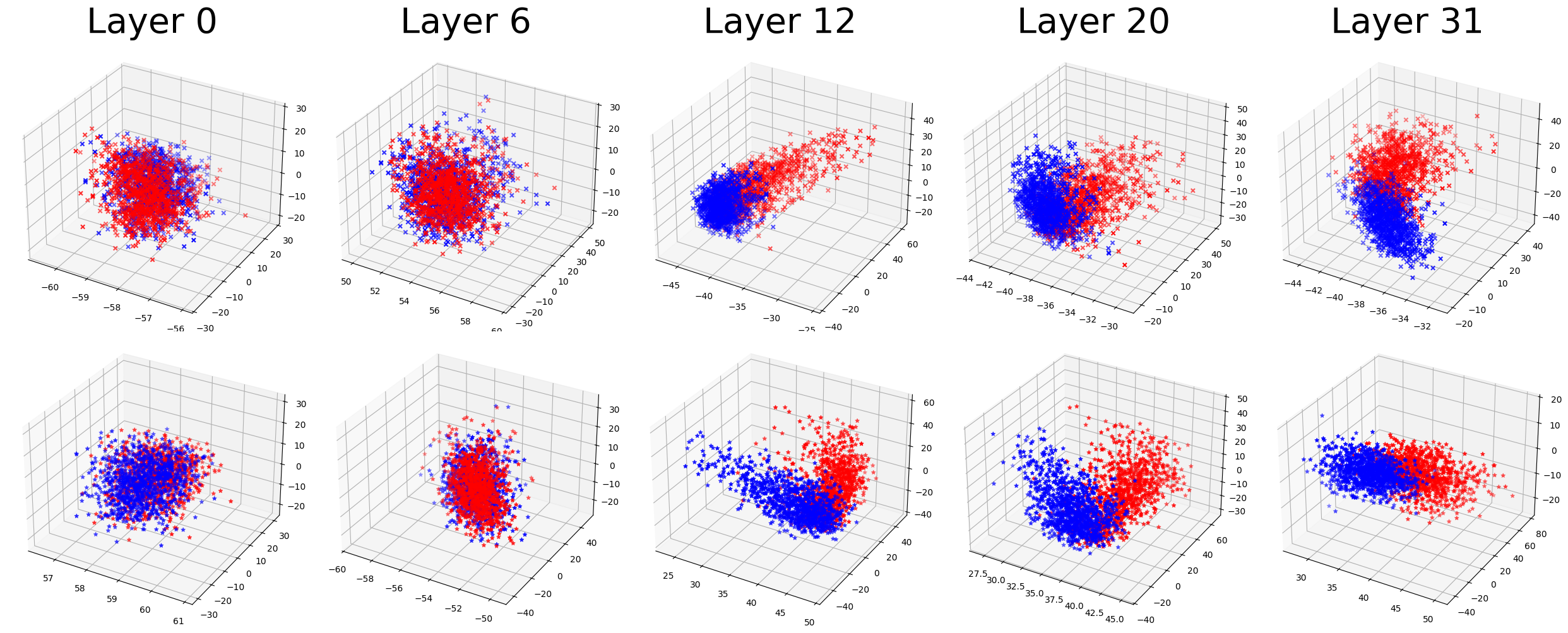}
        \caption{
    \textbf{Visualization of feature representations in LLaVA-1.5-7b}, reduced via PCA, shows red (associative) and blue (non-associative) points. The feature distributions show increasing separation in deeper layers, illustrating how associative distinctions are formed. See 
    Appendix ~\ref{sec_appendix:vis_fea_diff} 
    for all layers.
    }
        \label{fig:fea_vis_pca}   
    \end{minipage}
    \vspace{-0.5em}
\end{figure}

As shown in \Cref{fig:feature_diff_1,fig:feature_diff_2}, both cosine and Euclidean distances remain consistently low in the \textbf{shallow layers} (layers 0–9), indicating shared low-level perception. 
However, for middle and deep layers, we observe distinct patterns between cosine and Euclidean distance when comparing grounded and hallucinated responses across layers. Cosine distance peaks in the \textbf{middle layers} (layers 10–15), indicating that this stage is where feature directions diverge most significantly—suggesting that associative behavior is primarily introduced and shaped in this range. In contrast, Euclidean distance increases steadily across both \textbf{middle and deep layers} (layers 10–31), implying that the overall feature magnitudes continue to drift even in later stages. This discrepancy raises \textbf{\textcolor{red}{a key question:}} \textit{Is associative behavior actively introduced in the \textbf{deep layers}, or are these differences merely the propagated result of associative shifts originating in the \textbf{middle layers}?}

\begin{wrapfigure}{r}{0.48\linewidth}
    \centering  
    \includegraphics[width=\linewidth]{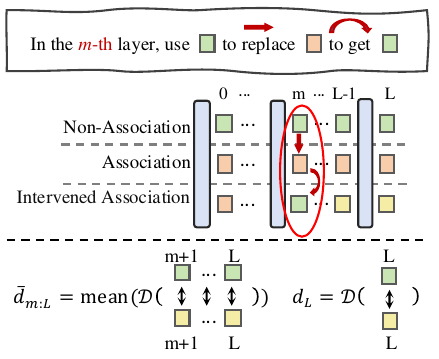}
    \caption{\textbf{Layer Intervention for Association Localization.} The goal is that locating the key layers for associative feature generation. Associative features are replaced with non-associative ones at different layers, and the impact on subsequent layers is evaluated using $d_L$ and $\bar{d}_{m:L}$.}
    \label{fig:find_layer}
    \vspace{-6em}
\end{wrapfigure}
\paragraph{Layer Intervention: Verifying the source of associative signals.}  
To answer this, we conduct a layer intervention experiment (\Cref{fig:find_layer}), in which we replace the associative feature $f^{(a)}_m$ with the corresponding non-associative feature $f^{(n)}_m$ at different layers $m$, and observe the influence on downstream representations. The modified feature propagation is defined as:
\begin{equation}
f^{\text{modified}}_l = \begin{cases}
f^{(a)}_l &  l < m \\
\mathcal{M}^l \circ \cdots \circ \mathcal{M}^{m+1}(f^{(n)}_m) &  l \geq m,
\end{cases}
\end{equation}
where $ \mathcal{M}^l $ denotes the $ l $-th layer of the model. We evaluate the impact by calculating the final layer difference $ d_L $ and the average layer difference $ \bar{d}_{m+1:L} $ as follows:
\begin{align}
    d_L &= \mathcal{D}(f^{\text{modified}}_L, f^{(n)}_L)\\
    \bar{d}_{m:L} &= \frac{1}{L - m} \sum_{i = m+1}^{L} \mathcal{D}(f^{\text{modified}}_i, f^{(n)}_i),
\end{align}
where $ \mathcal{D}(\cdot) $ denotes either cosine or Euclidean distance.

Results in \Cref{fig:feature_diff_3,fig:feature_diff_4} show that replacing features in shallow layers (layers 0-9) leads to minimal changes in downstream representations, indicating limited influence on associative processing. In contrast, replacing features in middle layers (layers 10-15) significantly reduces divergence in later layers, suggesting that these layers are the primary source of associative behavior. Replacements in deep layers (layers 16-31) again have limited impact, implying that these layers mainly propagate rather than generate associative features. More visualization in 
Appendix ~\ref{sec_appendix:vis_layer_intervention}.

\vspace{-0.2em}
\tcbset{width=\textwidth, colback=blue!5!white, colframe=blue!75!black}
\begin{tcolorbox}
\vspace{-0.2em}
    \textbf{Finding 1}: Middle layers are critical for shaping MLLM’s associative behavior.
\vspace{-0.2em}
\end{tcolorbox}
\vspace{-0.2em}

\subsection{Analyzing control strategies for associative behavior modulation}
\label{sec:analyze_control_strategy}

This analysis investigates whether associative behavior can be modulated by manipulating middle-layer representations, and whether hallucinated responses reveal effective directions for such control. Using the same grounded and hallucinated feature pairs from \Cref{sec:analyze-layer-localization}, we compute feature differences layer by layer to derive the control direction:
\begin{equation}
    v_l = f^{(a)}_l - F^{(n)}_l.
\end{equation}
We then apply this steering vector during inference to modulate the model’s output by adjusting the middle-layer features with control coefficient $\alpha$:
\begin{equation}
    f^{\text{control}}_l = f_l + \alpha \cdot v_l.
\end{equation}
To assess the impact of steering on associative behavior, we introduce VDAT (Visual-Divergent Association Test), a benchmark that evaluates a model's associative reasoning by prompting it to generate unrelated nouns to the input image, thereby measuring its capacity for visual-driven divergent thinking (details in \Cref{sec:experimental_setup}). 
As shown in \Cref{fig:factor}, increasing $\alpha$ from -1.5 to 1.5 raises CHAIR from approximately 38.8 to 53.6 and VDAT from around 83 to 87.9, indicating that higher $\alpha$ values lead to both more hallucination and stronger associative ability. Conversely, decreasing $\alpha$ reduces both scores. These results highlight that $\alpha$ provides a controllable mechanism for modulating associative behavior in MLLMs. 
These results yield two key findings:
\vspace{-0.5em}
\tcbset{width=\textwidth, colback=blue!5!white, colframe=blue!75!black}
\begin{tcolorbox}
\vspace{-0.2em}
    \textbf{Finding 2}: Modifying middle layers enables control over associative reasoning strength.\\
    \textbf{Finding 3}: Hallucinations help derive steering vectors to guide associative reasoning.
\vspace{-0.2em}
\end{tcolorbox}

\begin{figure*}[t]
    \begin{center}
    \includegraphics[width=\textwidth]{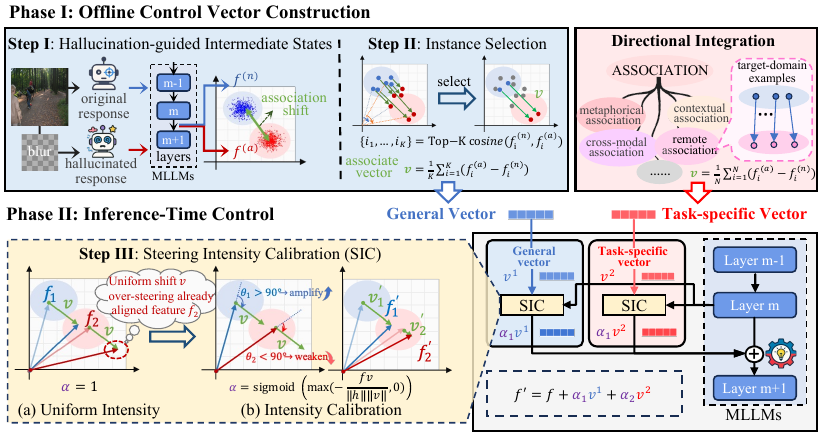}
    \end{center}
    \caption{\textbf{Overview of the proposed FlexAC framework.} 
    \textbf{Phase I: Offline Control Vector Construction} extracts a {\textbf{general associative vector}} from hallucination-guided intermediate features (Step I), by selecting Top-K instance pairs with maximal association shifts (Step II). It also constructs {\textbf{task-specific associative vectors}} from a few target-domain examples (Step III), reflecting diverse associative needs. 
    \textbf{Phase II: Inference-Time Control} injects these vectors into middle-layer features. A Steering Intensity Calibration (SIC) module adaptively adjusts the influence of each vector per sample to achieve controllable associative reasoning strength.
    }
    \label{fig:pipeline}
    \vspace{-1em}
\end{figure*}

\subsection{Flexible association control}
\label{sec:associative_manipulation}

Based on our findings in \Cref{sec:analyze-layer-localization,sec:analyze_control_strategy}, we propose \textbf{Flexible Association Control (FlexAC)}, a lightweight, training-free framework for modulating associative behavior in MLLMs. As illustrated in \Cref{fig:pipeline}, FlexAC operates in two phases: (I) \textbf{Offline Control Vector Construction}, which derives general and task-specific associative directions, and (II) \textbf{Inference-Time Control}, which injects these directions into middle-layer features for dynamic modulation.

\paragraph{Phase I: Offline Control Vector Construction.} 
To capture a general associative direction, we first induce hallucinated responses that exhibit high associative behavior (\textbf{Finding 3}). For each input, we extract hidden features from the middle layer $l$, where associative distinctions are most prominent (\textbf{Finding 1}), resulting in paired features $f_l^{(a)}$ and $f_l^{(n)}$. We select the  top-$K$ pairs with the highest cosine distances to construct a representative direction vector:
\begin{align}
\mathcal{I} &= \mathrm{Top-K} \left( \mathcal{D}_{\text{cos}}(f^{(a)}_{l,i}, f^{(n)}_{l,i}) \right);   
  v_l = \frac{1}{|\mathcal{I}|} \sum_{i \in \mathcal{I}} \left( f^{(a)}_{l,i} - f^{(n)}_{l,i} \right)
\end{align}

To handle tasks requiring diverse associative patterns (\eg, metaphorical, contextual), we further construct task-specific associative vectors from a few high-association, instruction-aligned examples. As vanilla MLLMs struggle to produce such outputs, we leverage GPT-4o to generate high-quality associative outputs.

\paragraph{Phase II: Inference-Time Control.}
During inference-time phase, we adjust the hidden state $f_l$ at middle layer $l$ (\textbf{Finding 2}) by injecting a combination of general associative vector $v^{\text{gen}}_l$ and task-specific associative vector $v^{\text{task}}_l$:
\begin{equation}
    f^{\text{control}}_l = f_l + \alpha_{\text{gen}} \cdot v^{\text{gen}}_l + \alpha_{\text{task}} \cdot v^{\text{task}}_l 
\end{equation}
where $\alpha$ is the tunable coefficient that controls the steering intensity. 
This formulation is grounded in recent theoretical findings~\cite{Li2025WhenIT}, which reveal that task-specific differences in model weights exhibit linearly decomposable structures. This property supports our assumption that associative directions can be independently extracted and combined within the hidden space.

However, directly applying a uniform steering vector across all inputs can lead to over-steering, especially when the input already exhibits strong associative behavior, causing deviation from the intended semantic space (see Step III of \Cref{fig:pipeline}). To mitigate it, we introduce steering intensity calibration strategy, which adjusts the steering strength $\alpha$ based on:
\begin{equation}
    \alpha = \mathrm{sigmoid} \left( \max \left( -\frac{f_l \cdot v_l}{\|f_l\|\|v_l\|}, 0 \right) \right)
\end{equation}
This formulation increase steering strength when the current representations is misaligned with the associate direction, and suppresses it when already aligned. We further normalize the modulated feature to preserve its scale:
\begin{equation}
    f^{\text{control}}_l \leftarrow f^{\text{control}}_l \cdot \frac{\|f_l\|}{\|f^{\text{control}}_l\|}
\end{equation}
This mechanism enables precise, interpretable modulation of associative behavior, allowing MLLMs to shift smoothly between factual accuracy and creative generation (\Cref{fig:vis_creativity}).

\section{Experiments}

\subsection{Experimental setup}
\label{sec:experimental_setup}

\noindent\textbf{Evaluation Metric}:
To evaluate the effectiveness of FlexAC, we conduct experiments on three benchmark types: (1) \textbf{hallucination}, using CHAIR~\cite{rohrbach2018CHAIR} and POPE~\cite{li2023POPE} to assess object-level factual consistency; (2) \textbf{creativity}, using our proposed VDAT for associative reasoning and Creation-MMBench~\cite{fang2025creation_bench} for open-ended image-grounded generation; and (3) \textbf{general-purpose capability}, using MME~\cite{fu2023mme}, MMMU~\cite{yue2023mmmu} and MMStar~\cite{chen2024mmstar} to ensure core perception and reasoning are preserved. Metric details are in 
Appendix ~\ref{sec_appendix:metrics_details}.

\begin{wrapfigure}{r}{0.5\linewidth}
    \centering  
    \includegraphics[width=0.8\linewidth]{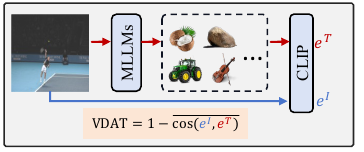}
    \caption{
    \textbf{Visual Divergent Association Test (VDAT)} evaluates a model’s associative reasoning by prompting it to generate unrelated nouns from an image, and quantifies performance through image-text measured using CLIP embeddings. 
    }
    \label{fig:VDAT}
    \vspace{-1em}
\end{wrapfigure}

\noindent\textit{\textbf{VDAT}: Visual Divergent Association Test.} 
To measure a model’s associative reasoning and creative potential more directly, we introduce VDAT, a diagnostic benchmark that complements Creation-MMBench by focusing specifically on associative reasoning strength. Inspired by \cite{chen2023DAT_LLM}, VDAT prompts the model to generate multiple nouns that are unrelated both to the input image, capturing its capacity for visual-driven divergent thinking (\Cref{fig:VDAT}). The metric is computed using CLIP ViT-L/14 embeddings.

\noindent\textbf{Implementation Details.}
We evaluate the effectiveness of our FlexAC on LLaVA-1.5~\cite{liu2024improvedllava}, Qwen-VL~\cite{bai2023qwenvl}, and Deepseek-VL~\cite{wu2024deepseekvl2mixtureofexpertsvisionlanguagemodels}, comparing it with Ha-DPO~\cite{zhao2023hallucinations}, VCD~\cite{leng2024vcd}, and VAF~\cite{yin2025clearsight}.
From the COCO2014~\cite{lin2014mscoco} dataset, we randomly selected 2000 images and then applied Instance Selection 
to choose 50 images for generating the general association vector.
For the layer intervention, we manipulated the following layers based on each model's associative strength: Qwen-VL (layers 15, 16, 17), LLaVA-1.5 (layers 11, 12, 13), and Deepseek-VL (layers 4, 5, 6). For FlexAC-P (faithfulness-enhanced) and FlexAC-C (creativity-enhanced), the control coefficient $\alpha$ is set to -1 and 1, respectively. All experiments were conducted on 8×RTX 4090 GPUs. 
The parameter analysis of the number of images is provided in 
Appendix ~\ref{sec_appendix:ablation_sample_num}. 

\begin{table*}[ht]
    \caption{\textbf{Performance on hallucination benchmarks.} 
    FlexAC here denotes the version configured to suppress associative behavior, aiming to improve factual accuracy (faithfulness). 
    }
    \label{exp:hallu_vs_creativity}
    \begin{center}
        \resizebox{ \linewidth}{!}{
        \begin{tabular}{l l ccc ccc cc}

    \toprule
    \multirow{2}{*}{\bf Models} &\multirow{2}{*}{\bf Methods}  &\multicolumn{4}{c}{\bf CHAIR}  &\multicolumn{4}{c}{\bf POPE} \\
    \cmidrule(lr){3-6} \cmidrule(lr){7-10}
     & &\multicolumn{1}{c}{$\text{CHAIR}_{S}\downarrow$}  &\multicolumn{1}{c}{$\text{CHAIR}_{I}\downarrow$}  &\multicolumn{1}{c}{Recall}  &\multicolumn{1}{c}{Len}  &\multicolumn{1}{c}{F1-score}$\uparrow$ &\multicolumn{1}{c}{$\text{Accuracy}\uparrow$}  &\multicolumn{1}{c}{$\text{Precision}\uparrow$}   &\multicolumn{1}{c}{Recall}\\
    \midrule

    \multirow{4}{*}{Qwen-VL} 
        & Regular & 40.6 & 12.5 & 71.7 & 94.6 & 85.6 & 86.6 & \textbf{92.9} & 79.3 \\ 
        & VCD & 42.0 & 11.2 & 71.7 & 91.2 & 86.3 & 87.2 & 92.4 & 81.0 \\ 
        & VAF & 38.0 & 11.7 & 72.2 & 91.4 & 86.5 & 87.2 & 91.4 & 82.0 \\ 
        \cmidrule(lr){2-10}
        \rowcolor{gray!20}
        & FlexAC (Ours) & \textbf{19.2} & \textbf{5.4} & 62.5 & 74.8 & \textbf{87.1} & \textbf{87.4} & 89.3 &85.1 \\ 
        \midrule

    \multirow{5}{*}{LLaVA-1.5}
        & Regular & 50.8 & 14.3 & 79.7 & 97.3 & 86.5 & 87.2 & 91.5 & 82.0 \\ 
        & Ha-DPO & 36.8 & \textbf{10.4} & 74.0 & 88.3 & 83.9 & 85.3 & \textbf{92.6} & 76.7 \\ 
        & VCD & 51.0 & 15.5 & 79.1 & 98.9 & 84.3 & 84.9 & 88.1 & 80.7 \\ 
        & VAF & 47.8 & 13.7 & 79.2 & 96.1 & 86.9 & 87.1 & 87.9 & 85.9 \\ 
        \cmidrule(lr){2-10}
        \rowcolor{gray!20}
        & FlexAC (Ours) & \textbf{36.6} & \textbf{10.4} & 75.0 & 95.1 & \textbf{87.9} & \textbf{87.8} & 87.1 & 88.8 \\ 
        \midrule

    \multirow{4}{*}{Deepseek-VL2} 
        & Regular & 32.6 & 9.2 & 67.0 & 121.0 & 88.5 & 88.4 & 88.1 & 88.8 \\ 
        & VCD & 36.6 & 11.3 & 67.2 & 128.2 & 87.9 & 87.8 & 87.6 & 88.1 \\ 
        & VAF & 32.0 & 9.2 & 66.2 & 119.0 & 88.5 & 88.4 & 87.6 & 89.4 \\ 
        \cmidrule(lr){2-10}
        \rowcolor{gray!20}
        & FlexAC (Ours) & \textbf{28.6} & \textbf{8.1} & 64.7 & 117.0 & \textbf{88.6} & \textbf{88.5} & \textbf{88.4} & 88.7 \\ 

    \bottomrule
    \end{tabular}
        }
    \end{center}
    \vspace{-2em}
\end{table*}

\subsection{Main results}

\paragraph{Results on Hallucination Benchmark.}
To evaluate FlexAC’s ability to improve factual accuracy in faithfulness-focused tasks, we conduct experiments on CHAIR and POPE. 
To this end, we set $\alpha$ in FlexAC to 1, selecting the precision-optimized variant.
As shown in ~\Cref{exp:hallu_vs_creativity}, FlexAC consistently achieves the lowest hallucination scores on most models and metrics. 
For examples, on CHAIR\textsubscript{S}, FlexAC reduces hallucination to 19.2 ($\uparrow$21.4) on Qwen-VL, 36.6 ($\uparrow$14.2 vs. Regular) on LLaVA-1.5, and 28.6 ($\uparrow$4.0) on Deepseek-VL2. On CHAIR\textsubscript{I}, it similarly achieves the best scores (5.4, 10.4, and 8.1 respectively). In terms of POPE accuracy, FlexAC achieves the highest F1-score on LLaVA-1.5 (87.9) and comparable or superior precision and recall across the board.
These results highlight FlexAC’s ability to flexibly suppress excessive associative behavior in factual tasks, leading to improved accuracy across models.

\begin{wraptable}{r}{0.6\linewidth}
    \vspace{-1em}
    \caption{\textbf{Performance on VDAT.} FlexAC here denotes the version optimized to enhance associative behavior for creative tasks (creativity). }
    \label{exp:vdat}
    \centering
    \resizebox{ \linewidth}{!}{
    \begin{tabular}{l ccc}

    \toprule
    \multirow{1}{*}{\bf Methods}  &\multicolumn{1}{c}{\bf Qwen-VL}  &\multicolumn{1}{c}{\bf LLaVA-1.5}  &\multicolumn{1}{c}{\bf DeepSeek-VL2} \\

    \midrule
         Regular & 84.85 & 86.89 & 84.54 \\ 
         Ha-DPO & - & 85.11 & - \\ 
         VCD & 83.69 & 86.83 & 84.62 \\ 
         VAF & 84.95 & 86.79 & 84.61 \\ 
        \midrule
        \rowcolor{gray!20}
         FlexAC (Ours) & \textbf{86.58} & \textbf{88.49} & \textbf{84.76} \\ 
         
    \bottomrule
    \end{tabular}
    }
    \vspace{-1.5em}
\end{wraptable}
\paragraph{Results on Creativity Benchmark.}
To evaluate FlexAC’s ability to enhance associative reasoning in creative tasks, we conduct experiments on VDAT(\Cref{exp:vdat}) and Creation-MMBench(\Cref{exp:creation_bench}).

As shown in \Cref{exp:vdat}, hallucination mitigation methods like Ha-DPO reduce hallucinations but impair associative capacity, leading to lower creativity (\eg, VDAT score of 85.11 vs. 86.89 for the regular model). 
In contrast, FlexAC improves remote associative reasoning, achieving a higher VDAT score of 88.49. 
To further verify the validity of the VDAT metric, we conduct a user study presented in 
Appendix ~\ref{sec_appendix:vdat_user}. 
Further, on Creation-MMBench \Cref{exp:creation_bench}, we report VFS (Visual Fidelity Score), which evaluates image-text alignment, and Reward, which quantifies creativity improvements relative to the base model (Qwen-VL). 
FlexAC achieves the highest Reward (10.92), outperforming methods like VCD (-3.86) and VAF (-1.63), while maintaining competitive VFS.

\begin{figure}
    \centering
    \includegraphics[width=0.9\textwidth]{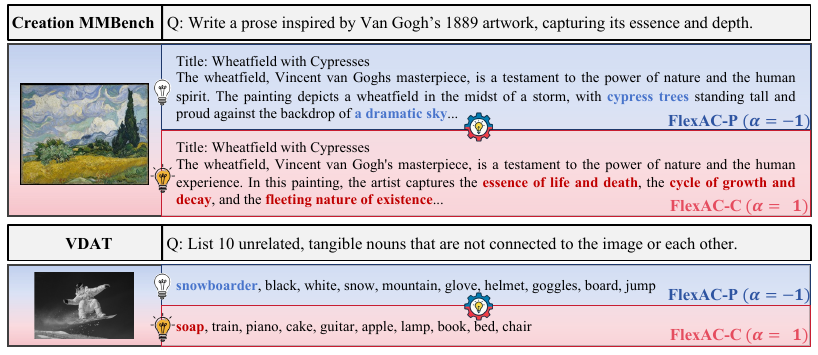}
    \caption{\textbf{Visualization of FlexAC’s Control on Associative Reasoning.} 
    This figure illustrates example outputs from Creation MMBench and VDAT, showing how FlexAC-P (faithfulness) and FlexAC-C (creativity) adjust the level of associative strength in the generated responses.
    }
    \label{fig:vis_creativity}
    \vspace{-2em}
\end{figure}

Qualitative examples in \Cref{fig:vis_creativity} further support this: in Creation-MMBench, FlexAC-P focuses on concrete visual elements (e.g., “cypress trees”), while FlexAC-C introduces abstract themes (e.g., “life and death”). In VDAT, FlexAC-P outputs image-relevant nouns (e.g., “snowboarder”), whereas FlexAC-C generates semantically distant words (e.g., “guitar”, “apple”), demonstrating enhanced divergent thinking.
These examples confirm that FlexAC effectively modulates associative strength to meet diverse creative demands. For additional examples, see 
Appendix ~\ref{sec_appendix:more_vis_examples}.

\begin{table}[ht]
    \vspace{-0.5em}
    \caption{\textbf{Performance on Creation-MMBench.} We report results on four subcategories: Literary Writing (LW), Common Functional Writing (CFW), Professional Functional Writing (PFW), and Creative Multimodal Understanding (CMU). FlexAC here denotes the version optimized to enhance associative behavior for creative tasks (creativity).}

    \label{exp:creation_bench}
    \begin{center}
    \resizebox{ 0.9\linewidth}{!}{
    \begin{tabular}{l ccccc ccccc}
    \toprule
        \multirow{2}{*}{\textbf{Methods}} & \multicolumn{2}{c|}{\textbf{Overall}} & \multicolumn{2}{c|}{\textbf{LW}} & \multicolumn{2}{c|}{\textbf{CMU}} & \multicolumn{2}{c|}{\textbf{PFW} } & \multicolumn{2}{c}{\textbf{CFW}} \\
    \cmidrule(lr){2-3}  \cmidrule(lr){4-5} \cmidrule(lr){6-7} \cmidrule(lr){8-9} \cmidrule(lr){10-11}
      & VFS & Reward & VFS & Reward & VFS & Reward & VFS & Reward & VFS & Reward \\
    \midrule
         Regular & 6.10 & 0.00 & 6.83 & 0.00 & 5.53 & 0.00 & 5.58 & 0.00 & 6.66 & 0.00 \\ 
         VCD & 6.05 & -3.86 & 6.68 & -2.71 & 5.67 & 2.50 & \textbf{5.61} & -3.77 & 6.46 & -6.57 \\ 
         VAF & 6.06 & -1.63 & 6.39 & -3.96 & 5.57 & -4.17 & \textbf{5.61} & -0.53 & 6.64 & -0.93 \\ 
        \midrule
        \rowcolor{gray!20}
         FlexAC (Ours) & \textbf{6.25}  &  \textbf{10.92} &  \textbf{7.20} &  \textbf{15.63} &   \textbf{5.83}   &  \textbf{6.11}  &  5.43    & \textbf{5.96} &   \textbf{7.00}  &  \textbf{15.65}  \\ 
    \bottomrule
    \end{tabular}
    }
    \end{center}
    \vspace{-1em}
\end{table}

\paragraph{Results on General-Purpose Benchmark.} 
To evaluate the generalization capabilities of FlexAC across a range of tasks, we conduct experiments on three standard multimodal benchmarks using Qwen-VL: MME, MMMU, and MMStar. These benchmarks cover a wide range of capabilities including fine-grained grounding, reasoning, and instruction following.

\begin{figure}[htb]
    \vspace{-1em}
    \centering
    \includegraphics[width=\textwidth]{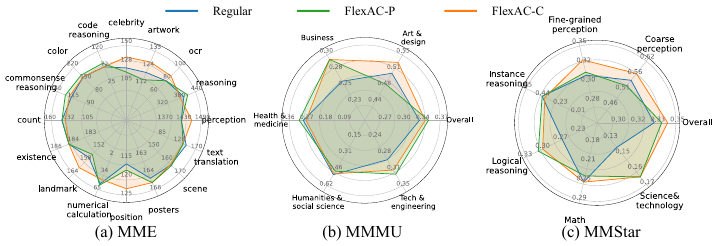}
    \caption{\textbf{Performance on general-purpose benchmarks.} Comparison of Regular, FlexAC-P (faithfulness-enhanced, $\alpha = -1$), and FlexAC-C (creativity-enhanced, $\alpha = 1$).}
    \label{fig:general_bench}
    \vspace{-1.5em}

\end{figure}

As shown in ~\Cref{fig:general_bench}, both FlexAC-P (faithfulness-enhanced) and FlexAC-C (creativity-enhanced) maintain performance similar to the vanilla model across most categories, indicating no significant compromise in general capabilities. 
Notably, FlexAC-C outperforms the baseline on the OCR task in MME, likely due to its enhanced ability to associate text with related visual entities, improving inference and disambiguation under challenging conditions.

\vspace{-0.5em}
\subsection{Ablation study}

\paragraph{Layer-wise Control Analysis.}
We investigate the impact of middle layers on associative reasoning and identify the optimal control layers by testing interventions on shallow, middle, and deep layers, evaluating their effects on both CHAIR and VDAT metrics.

\begin{figure*}[ht]
    \centering
    \includegraphics[width=0.9\textwidth]{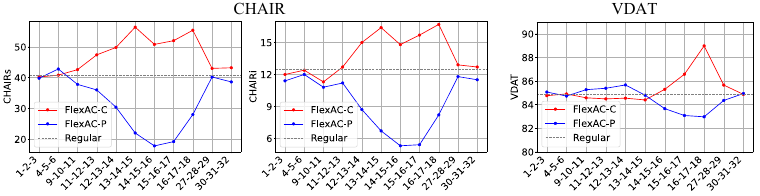}
    \caption{\textbf{Layer-wise analysis of control effectiveness in FlexAC.} The x-axis represents the control layers, while the y-axis shows the performance of the model on CHAIR and VDAT metrics.}
    \label{fig:ablation_layers}
    \vspace{-1em}
\end{figure*}

The results in \Cref{fig:ablation_layers} demonstrate that middle layers have the most significant impact on performance: FlexAC-P achieves the best CHAIR results when suppressing associative behavior, while FlexAC-C shows the highest VDAT scores when enhancing creativity. In contrast, controlling shallow or deep layers has minimal effect. Based on these findings, we select layers 15, 16, and 17 as the control layers for Qwen-VL; results for other models are provided in 
\Cref{sec_appendix:ablation_layer}.

\paragraph{Effectiveness of different Components.} 
We conducted an ablation study to assess the impact of components within FlexAC, including Instance Selection (IS), Steering Intensity Calibration (SIC), and Directional Integration (DI), on faithfulness (CHAIR) and creativity (VDAT).

\begin{wrapfigure}{r}{0.48\textwidth}
    \centering
    \includegraphics[width=\linewidth]{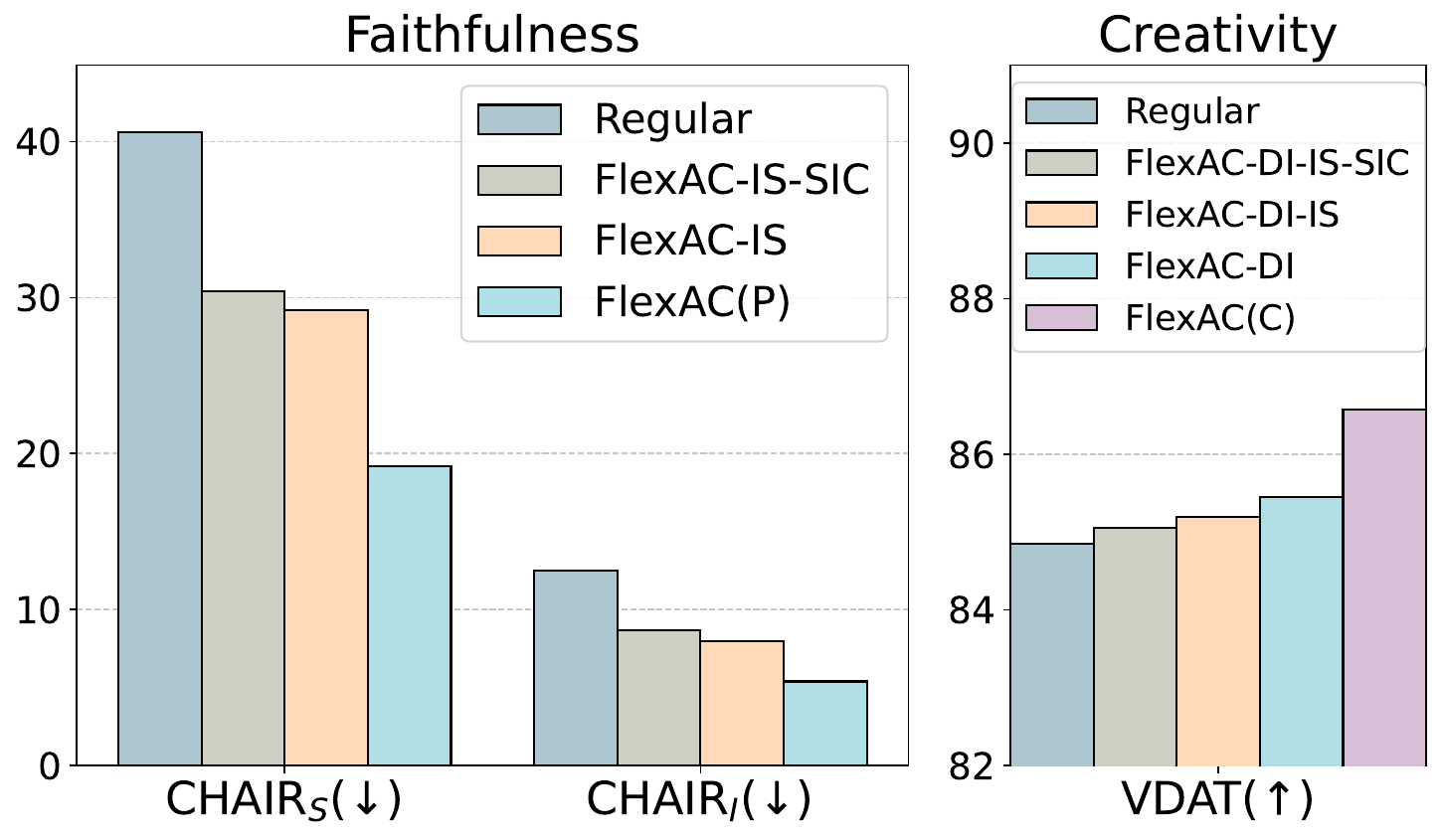}
    \caption{\textbf{Ablation study on components}, showing the impact of Instance Selection (IS), Steering Intensity Calibration (SIC), and Directional Integration (DI).
    }
    \label{fig:ablation_module}
    \vspace{-1em}
\end{wrapfigure}
As shown in \Cref{fig:ablation_module}, for CHAIR, FlexAC(P) achieves the lowest CHAIR\textsubscript{S} score (19.2), indicating effective hallucination reduction compared to the regular model (40.6). 
When IS and SIC are removed from FlexAC (FlexAC-IS-SIC), performance slightly worsens (30.4), confirming their role in enhancing faithfulness. 
Similarly, for creativity, FlexAC-C scores the highest VDAT (86.58). Removing IS and SIC in FlexAC-IS-SIC leads to a small decrease (85.05), while FlexAC-DI results in a slight improvement, highlighting the importance of DI for creativity.
In summary, FlexAC enables flexible adjustment of associative strength to meet the needs of different tasks, balancing hallucination reduction and creativity enhancement effectively.

\vspace{-1em}
\section{Related work}
\label{sec:related_work}
\vspace{-0.5em}

\noindent\textbf{Multimodal Large Language Models.}
Recent advances in large language models (LLMs)~\cite{gilardi2023chatgpt, touvron2023llama, bai2023qwenllm} have led to the emergence of multimodal LLMs (MLLMs) that incorporate visual inputs for enhanced capabilities~\cite{dai2023instructblip, li2022blip}. 
LLaVA~\cite{liu2023llava, liu2024improvedllava} improves instruction-following via visual instruction tuning, while Qwen-VL~\cite{bai2023qwenvl} enhances spatial reasoning through visual grounding. DeepSeek- dVL2~\cite{wu2024deepseekvl2mixtureofexpertsvisionlanguagemodels} adopts a Mixture-of-Experts architecture to improve multimodal comprehension.

\noindent\textbf{Hallucination in MLLMs.}
MLLMs still face various safety risks~\cite{yuan2022natural, sun2024targetoffenseadversarialexample, chen2025safeptr}. Hallucination is one of the core challenges, where MLLMs generate content misaligned with visual input~\cite{liu2024hallucination_survey}. To address this, VCD~\cite{leng2024vcd} employs contrastive decoding, and VAF~\cite{yin2025clearsight} enhances visual signal processing during fusion. HA-DPO~\cite{zhao2023hallucinations} reduces hallucinations via preference optimization.

\noindent\textbf{Creativity in Large Models.}
Creativity, involving divergent thinking and novel associations~\cite{guilford1967creativity, runco2012SDC}, has been explored in LLMs via cognitive theories like dual-pathway~\cite{beaty2014DAT}. Olson \etal~\cite{olson2021DAT} and Chen \& Ding~\cite{chen2023probing} promote remote associations; MacGyver~\cite{tian2024macgyver} and CLOT~\cite{zhong2024clot} tackle functional fixedness and divergent thinking, respectively. Creation-MMBench~\cite{fang2025creation_bench} provides a benchmark for evaluating image-grounded creative generation.

\vspace{-1em}
\section{Conclusion}

\vspace{-1em}
In this work, we investigate the root of associative behavior in MLLMs, finding that middle-layer representations govern associative reasoning strength and that hallucinated responses encode reliable steering directions. Based on these insights, we propose FlexAC, a lightweight, training-free framework that combines hallucination-guided steering with adaptive calibration and in-context augmentation. FlexAC enables controllable creativity and achieves state-of-the-art performance across hallucination, creativity, and general-purpose benchmarks. 
\textbf{Limitations:} FlexAC requires white-box access to hidden states and is not applicable to black-box models like ChatGPT.

\section{Acknowledgements}
This study is supported by grants from the National Natural Science Foundation of China (Grant No.
U23A20315, No. 62425208, No. U22A2097, No. 62122018, No. 62020106008), Shenzhen Science
and Technology Program (No.JCYJ20240813114208012), Fundamental Research Funds for the
Central Universities, and Natural Science Foundation of Sichuan Province (Grant No. 2025ZNSFSC1463).

\bibliographystyle{unsrt}
\bibliography{main_FlexAC_arxiv}



\resumetocwriting 

\appendix
\clearpage

\begin{center}
  \vspace*{2\baselineskip}
  \Large \textbf{{FlexAC}\raisebox{-0.18\height}{\includegraphics[height=1.1em]{figures/icon.png}}\xspace: Towards Flexible Control of Associative Reasoning in Multimodal Large Language Models \\
(Supplementary Material)}
\end{center}

\input{X_suppl.tex}

\end{document}

%% file: X_suppl.tex
{
    \hypersetup{
    linkcolor=black
}

{
\begin{minipage}{0.99\textwidth}
    \let\mtcontentsname\contentsname
    \renewcommand\contentsname{\MakeUppercase\mtcontentsname}
    \noindent
    \rule{\textwidth}{1.4pt}\\[-0.75em]
    \noindent
    \rule{\textwidth}{0.4pt}
    \tableofcontents
    \rule{\textwidth}{0.4pt}\\[-0.70em]
    \noindent
    \rule{\textwidth}{1.4pt}
  \end{minipage}\par
}

}
\clearpage

\section{Broader Impacts}

FlexAC introduces finer control over the associative behavior of MLLMs, enabling safer and more context-appropriate responses across tasks. This may benefit applications requiring factual precision (e.g., education, medical support) or creative output (e.g., storytelling, art generation). However, enhancing associative capacity also increases the model's expressive power, which—if misused—could lead to persuasive but unfounded generations. As with all generation-controlling techniques, FlexAC should be deployed alongside robust safeguards to ensure alignment with human intent and ethical use.

\section{Data Generation and Feature Extraction}
\label{exp_appendix:data_generate}

\begin{figure}[ht]
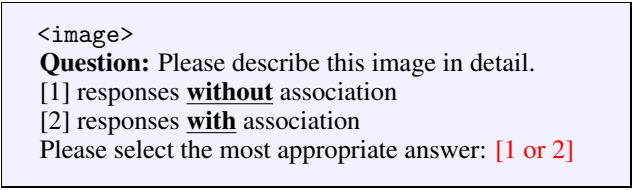

\centering
\begin{tcolorbox}[colback=blue!5!white, colframe=black, width=0.6\textwidth, sharp corners, boxrule=0.5pt]
\texttt{\textless image\textgreater} \\
\textbf{Question:} Please describe this image in detail. \\
{[1]} responses \textbf{\underline{without}} association \\
{[2]} responses \textbf{\underline{with}} association \\
Please select the most appropriate answer: \textcolor{red}{[1 or 2]}
\end{tcolorbox}
\caption{The prompt for extracting associative and non-associative features}
\label{fig_appendix:choice_prompt}
\end{figure}

\noindent\textbf{Inducing and Representing Model Associations}: To investigate the causes of model association, we generate two data distributions: one from the model's original outputs (non-associative) and another with induced associative content using blurred images and tailored prompts~\cite{leng2024vcd, wang2024mitigating}. For example, the model is prompted with: “Describe the image and include some hallucinated objects that are imagined but do not exist in the image, as if they were real.” Following \cite{rimsky2024caa}, we construct a multiple-choice dataset to capture feature distributions. The model is given an image and prompted to generate detailed responses, with two predefined options (\Cref{fig_appendix:choice_prompt}): [1] non-associative (factual) and [2] associative (creative). The hidden states corresponding to these inputs are extracted to obtain distinct feature representations, $F_{\text{non-assoc}}^l$ and $F_{\text{assoc}}^l$, capturing the model’s internal response to both associative and non-associative prompts across different layers. 

\section{Metrics details}
\label{sec_appendix:metrics_details}


All comparative experiments are conducted using the VLMEvalKit\footnote{\url{https://github.com/open-compass/VLMEvalKit}}. For binary choice questions, we prompt the model with: ``Please answer Yes or No.'' We evaluate three models in our experiments: LLaVA-1.5~(\texttt{liuhaotian/llava-v1.5-7b}), Qwen-VL~(\texttt{Qwen/Qwen-VL-Chat}), and DeepSeek-VL2~(\texttt{deepseek-ai/deepseek-vl2-tiny}).

\paragraph{VDAT}: VDAT fills a gap in evaluating the creative potential of multimodal models, which previous metrics did not adequately address. To ensure consistency, both CHAIR and VDAT were evaluated using the same 500 images, randomly selected from the MSCOCO dataset. 

\paragraph{Creation-MMBench}\cite{fang2025creaion-mmbench}: Creation-MMBench is a multimodal benchmark designed to evaluate the creative capabilities of MLLMs in real-world, image-grounded scenarios. It contains 765 test cases across 51 fine-grained tasks, with instance-specific criteria that assess both imaginative quality and visual consistency. In contrast to prior work that compares models to GPT-4o, our evaluation focuses on measuring improvements over each model’s own vanilla baseline.

\paragraph{CHAIR}\cite{rohrbach2018CHAIR}: Caption Hallucination Assessment with Image Relevance (CHAIR) is a metric designed to evaluate the hallucination of image caption task. It measures the hallucination rate of the generated text by comparing the generated caption with the ground-truth caption. CHAIR consists of two metrics: CHAIR$_{S}$ and CHAIR$_{I}$. They can be calculated as follows:
\begin{align}
    \text{CHAIR}_{S} &= \frac{|\{hallucinated\ \ objects\}|}{|\{all\ \ mentioned\ \ objects\}|}, \\
    \text{CHAIR}_{I} &= \frac{|\{captions\ \ with\ \ hallucinated objects\}|}{|\{all\ \ captions\}|}.
\end{align}

\paragraph{POPE}\cite{li2023POPE}: The Polling-based Object Probing Evaluation (POPE) is a metric developed to evaluate object hallucination in MLLMs. By framing the evaluation as a series of Yes-or-No questions about specific objects in images, POPE avoids issues related to instruction sensitivity. Using three sampling strategies—Random, Popular, and Adversarial—it effectively examines models’ tendencies to hallucinate frequently occurring or co-occurring objects, providing a stable and reliable assessment of object hallucination. Refer to \cite{li2023POPE}, we built POPE on 500 randomly selected MSCOCO~\cite{lin2014mscoco} validation images, each containing over three ground-truth objects and six constructed questions.

\paragraph{MME}\cite{fu2023mme}: MLLM Evaluation benchmark (MME) is a benchmark designed to assess multimodal large language models (MLLMs) across core skills in perception and cognition, such as object recognition, attribute identification, reasoning, and translation. Using accuracy-based metrics, MME provides objective insights into model capabilities, highlighting areas for improvement in understanding and reasoning. 

\paragraph{MMMU}\cite{yue2023mmmu}: MMMU (Massive Multi-discipline Multimodal Understanding and Reasoning) is a large-scale benchmark targeting expert-level multimodal understanding and reasoning. It comprises 11.5K college-level questions across 6 disciplines and 30 subjects, featuring 30 diverse image types such as charts, medical scans, diagrams, and chemical structures. MMMU emphasizes deep domain knowledge and deliberate reasoning, challenging models to integrate perception, knowledge, and logic in complex tasks. It serves as a necessary testbed for evaluating progress toward Expert AGI.

\paragraph{MMStar}\cite{chen2024mmstar}: MMStar is a high-quality benchmark designed to evaluate vision-language models on truly vision-dependent tasks. It includes 1,500 human-curated samples across 6 core capabilities and 18 fine-grained skills, ensuring minimal data leakage and strong visual grounding.

\clearpage
\section{Detailed Experimental Results}

\subsection{User study}
\label{sec_appendix:vdat_user}

To validate the effectiveness of the VDAT metric as a measure of associative creativity, we conducted a human evaluation study comparing FlexAC against several baselines. Specifically, we randomly selected 30 image-response examples from the Qwen-VL evaluation set and presented them to 15 human raters. For each example, two responses were shown—one from FlexAC and one from a baseline method (Regular, VAF, or VCD). Participants were asked to judge which response contained objects more unrelated to the image, as a proxy for stronger remote association. The response options were presented as “Answer A” and “Answer B,” with the method-to-label mapping randomized in each trial to eliminate bias. 
Raters evaluated each pair on a five-point scale ranging from “A is much better than B” to “B is much better than A.” These choices were then converted to numeric scores for aggregation—for example, ``$A >> B$'' assigns 3 points to A, ``$A = B$'' assigns 1 point to both A and B.

\begin{figure}[h]
    \centering
    \includegraphics[width=0.8\linewidth]{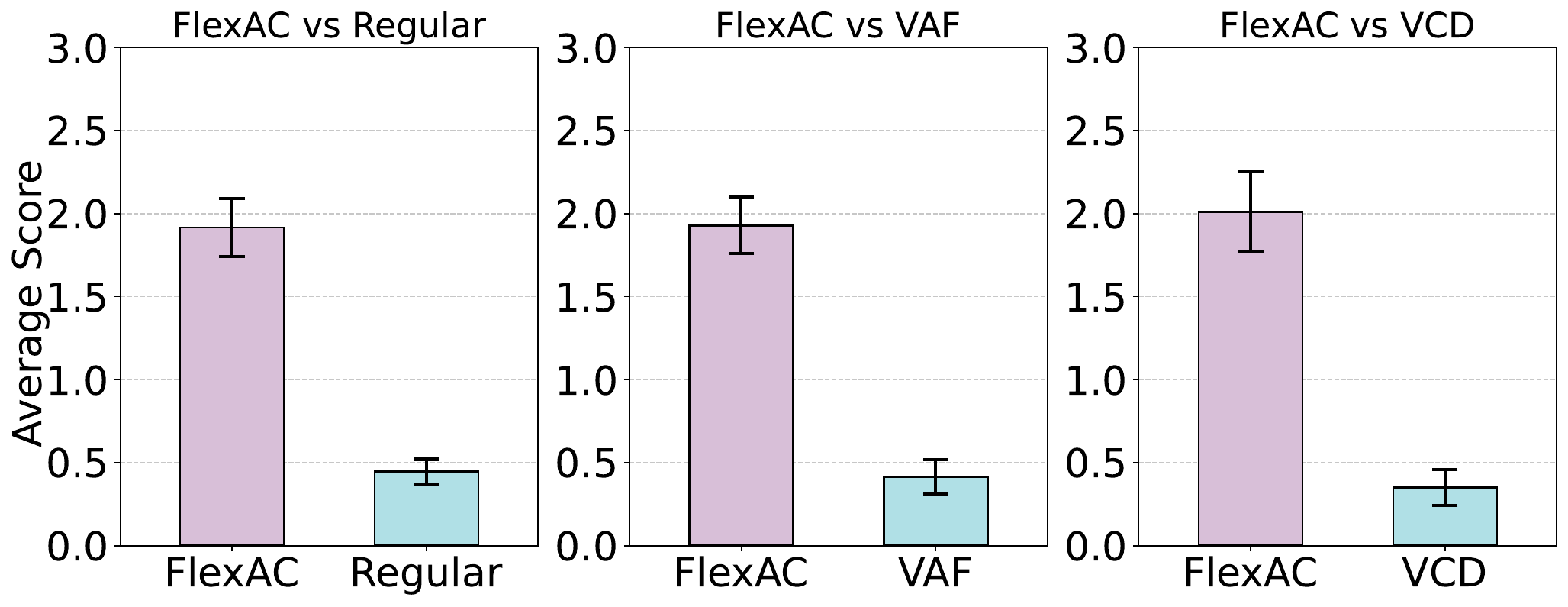}
    \caption{Average user ratings comparing FlexAC with baseline methods on the VDAT task. Each bar represents the average score across 15 users for 30 randomly selected image-response pairs. Error bars indicate the maximum and minimum individual user scores, reflecting rating consistency. Higher scores indicate stronger perceived remote association ability.}
    \label{fig_appendix:user_study_with_error_bars}
\end{figure}

\begin{figure}[h]
    \centering
    \includegraphics[width=0.7\linewidth]{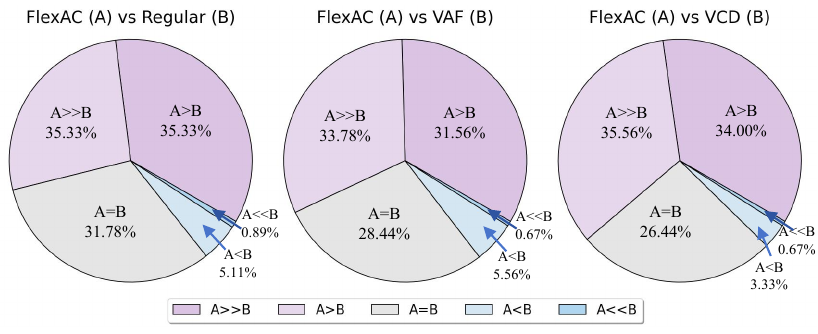}
    \caption{Distribution of user rating preferences when comparing FlexAC with each baseline on the VDAT task. $A = B$ indicates equal preference; $A >> B$ and $A > B$ mean FlexAC is preferred; $A << B$ and $A < B$ mean the baseline is preferred. Results show strong preference for FlexAC in most cases.}
    \label{fig_appendix:user_study_win_rate}
\end{figure}

\begin{figure}[h]
    \centering
    \includegraphics[width=\linewidth]{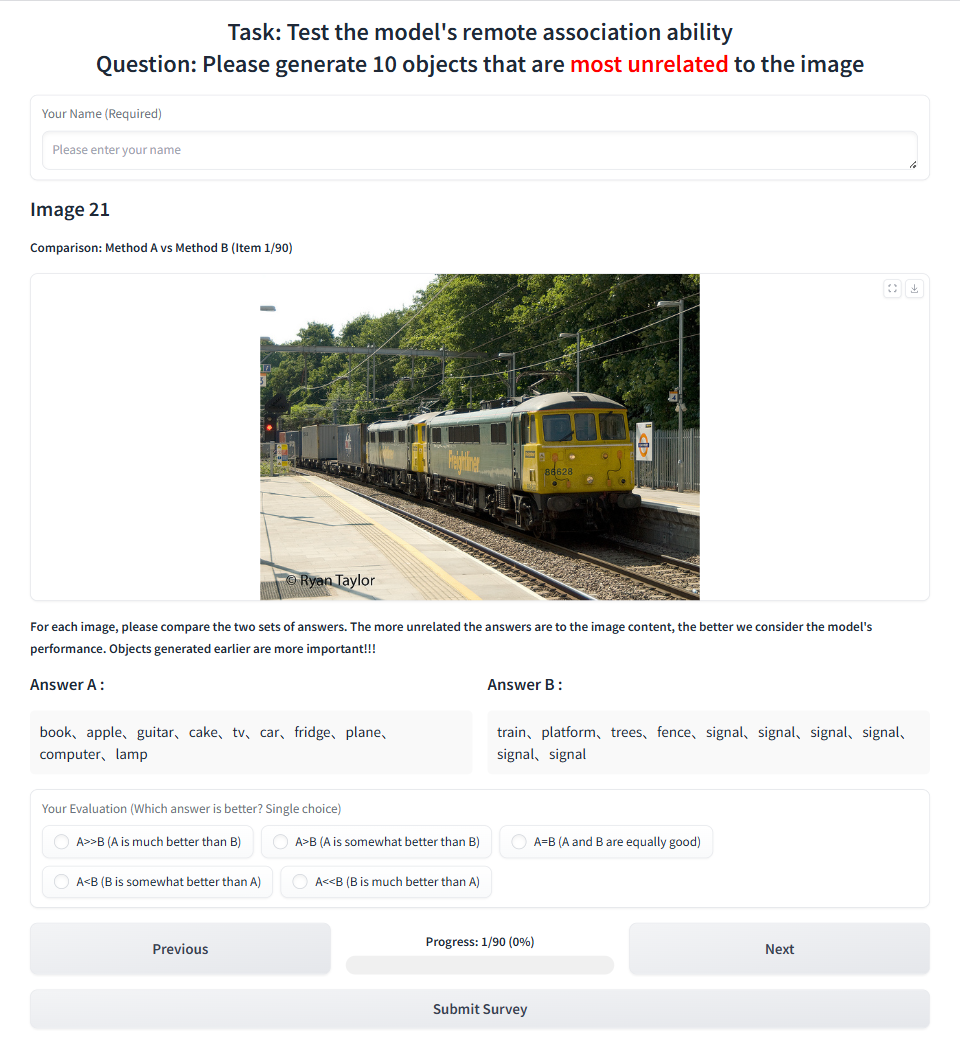}
    \caption{Interface of the user study for evaluating remote association. Participants are presented with an image and two model-generated answers, and asked to judge which set of objects is more unrelated to the image. The label A or B was randomly assigned to FlexAC or baseline in each trial to prevent method identification.}
    \label{fig_appendix:user_study_interface}
\end{figure}

\Cref{fig_appendix:user_study_with_error_bars} shows that FlexAC consistently receives higher average scores than all baselines, with low variance across users. \Cref{fig_appendix:user_study_win_rate} further reveals that over 70\% of responses favored FlexAC ($A > B$ or $A >> B$), while fewer than 6\% favored the baseline. These results demonstrate strong alignment between the VDAT metric and human judgment. \Cref{fig_appendix:user_study_interface} provides a screenshot of the evaluation interface. Together, these findings support VDAT as a valid and human-aligned metric for measuring associative creativity in vision-language generation tasks.


\clearpage
\subsection{Extended results on Creation-MMBench}
\label{sec_appendix:more_model_creation_mmbench}

To further evaluate FlexAC’s effectiveness in enhancing associative behavior for creative generation, we report additional results on the Creation-MMBench benchmark using two base models: LLaVA-1.5 and DeepSeek-VL2, as shown in \Cref{exp_appendix:more_modal_on_creation_bench}. This benchmark covers four creative subcategories—Literary Writing (LW), Common Functional Writing (CFW), Professional Functional Writing (PFW), and Creative Multimodal Understanding (CMU). For each subtask, we report two metrics: VFS (Visual Fidelity Score), which measures the alignment between the image and the generated response, and Reward, which quantifies creativity improvements relative to the base model (i.e., vanilla LLaVA-1.5 or vanilla DeepSeek-VL2, respectively).

In this experiment, FlexAC is configured to enhance associative behavior, with the goal of generating more creative content. Across both models, FlexAC achieves the highest overall reward scores, demonstrating its effectiveness in promoting creative generation without sacrificing visual grounding. Notably, on DeepSeek-VL2, FlexAC obtains a reward of +10.35 on PFW and +6.73 overall, clearly outperforming all baselines. To test whether performance gains stem from meaningful control or arbitrary perturbation, we also evaluate a variant that injects random vectors into the representation. As shown in the “Random” rows, this leads to large performance drops across all metrics—highlighting that FlexAC’s improvements do not come from noise or randomness, but from targeted modulation of associative features. These results further support FlexAC’s ability to improve creative reasoning across diverse multimodal architectures.

\begin{table*}[ht]
    \caption{\textbf{Performance on Creation-MMBench.} We report results on four subcategories: Literary Writing (LW), Common Functional Writing (CFW), Professional Functional Writing (PFW), and Creative Multimodal Understanding (CMU). FlexAC here denotes the version optimized to enhance associative behavior for creative tasks (creativity).}
    \label{exp_appendix:more_modal_on_creation_bench}
    \begin{center}
    \resizebox{ 0.9\linewidth}{!}{
    \begin{tabular}{ll ccccc ccccc}
    \toprule
    \multirow{2}{*}{\textbf{Models}} & \multirow{2}{*}{\textbf{Methods}} & \multicolumn{2}{c|}{\textbf{Overall}} & \multicolumn{2}{c|}{\textbf{LW}} & \multicolumn{2}{c|}{\textbf{CFW}} & \multicolumn{2}{c|}{\textbf{PFW}} & \multicolumn{2}{c}{\textbf{CMU}} \\
    \cmidrule(lr){3-4}  \cmidrule(lr){5-6} \cmidrule(lr){7-8} \cmidrule(lr){9-10} \cmidrule(lr){11-12}
     & & VFS & Reward & VFS & Reward & VFS & Reward & VFS & Reward & VFS & Reward \\
    \midrule

    \multirow{5}{*}{LLaVA1.5}
        & Regular & 5.32 & 0.00 & 6.28 & 0.00 & nan & 0.00 & 4.26 & 0.00 & 6.08 & 0.00 \\ 
        & Random & 3.53 & -60.49 & 3.11 & -69.58 & 2.19 & -72.22 & 2.93 & -60.35 & 4.80 & -52.69 \\ 
        & Ha-DPO & 4.84 & -26.41 & 5.09 & -30.00 & 3.68 & -19.72 & 4.37 & -26.23 & 5.67 & -27.22 \\ 
        & VCD & \textbf{5.56} & 2.00 & \textbf{6.69} & 7.08 & \textbf{4.87} & \textbf{5.00} & \textbf{4.86} & 3.00 & \textbf{6.23} & -2.31 \\ 
        & VAF & 5.30 & -5.86 & 6.15 & -3.54 & 4.27 & -5.00 & 4.74 & -6.34 & 6.01 & -6.67 \\ 
        \cmidrule(lr){2-12}
        \rowcolor{gray!20}
        & FlexAC (Ours) & 5.45 & \textbf{4.39} & 6.52 & \textbf{11.88} & 4.76 & -3.89 & 4.72 & \textbf{3.62} & 6.18 & \textbf{4.63} \\ 
        \midrule

    \multirow{4}{*}{DeepSeek-VL2}  
        & Regular & 6.12 & 0.00 & 6.98 & 0.00 & 6.35 & 0.00 & 5.71 & 0.00 & 6.21 & 0.00 \\ 
        & Random & 2.34 & -77.47 & 1.32 & -78.96 & 3.28 & -75.83 & 1.96 & -82.46 & 2.75 & -72.08 \\ 
        & VCD & \textbf{6.42} & 4.80 & \textbf{7.37} & \textbf{5.63} & \textbf{6.58} & -3.33 & 5.98 & 6.40 & \textbf{6.55} & 5.46 \\ 
        & VAF & 6.26 & -0.39 & 6.70 & -1.25 & 6.46 & -3.06 & 5.93 & 2.46 & 6.42 & -2.13 \\ 
        \cmidrule(lr){2-12}
        \rowcolor{gray!20}
        & FlexAC (Ours) & 6.29 & \textbf{6.73} & 6.76 & 0.63 & 6.37 & \textbf{4.17} & \textbf{5.99} & \textbf{10.35} & 6.44 & \textbf{6.48} \\ 
        
    \bottomrule
    \end{tabular}
    }
    \end{center}
\end{table*}

\clearpage
\subsection{Efficiency comparison}

\begin{figure}[h]
    \centering
    \includegraphics[width=0.45\linewidth]{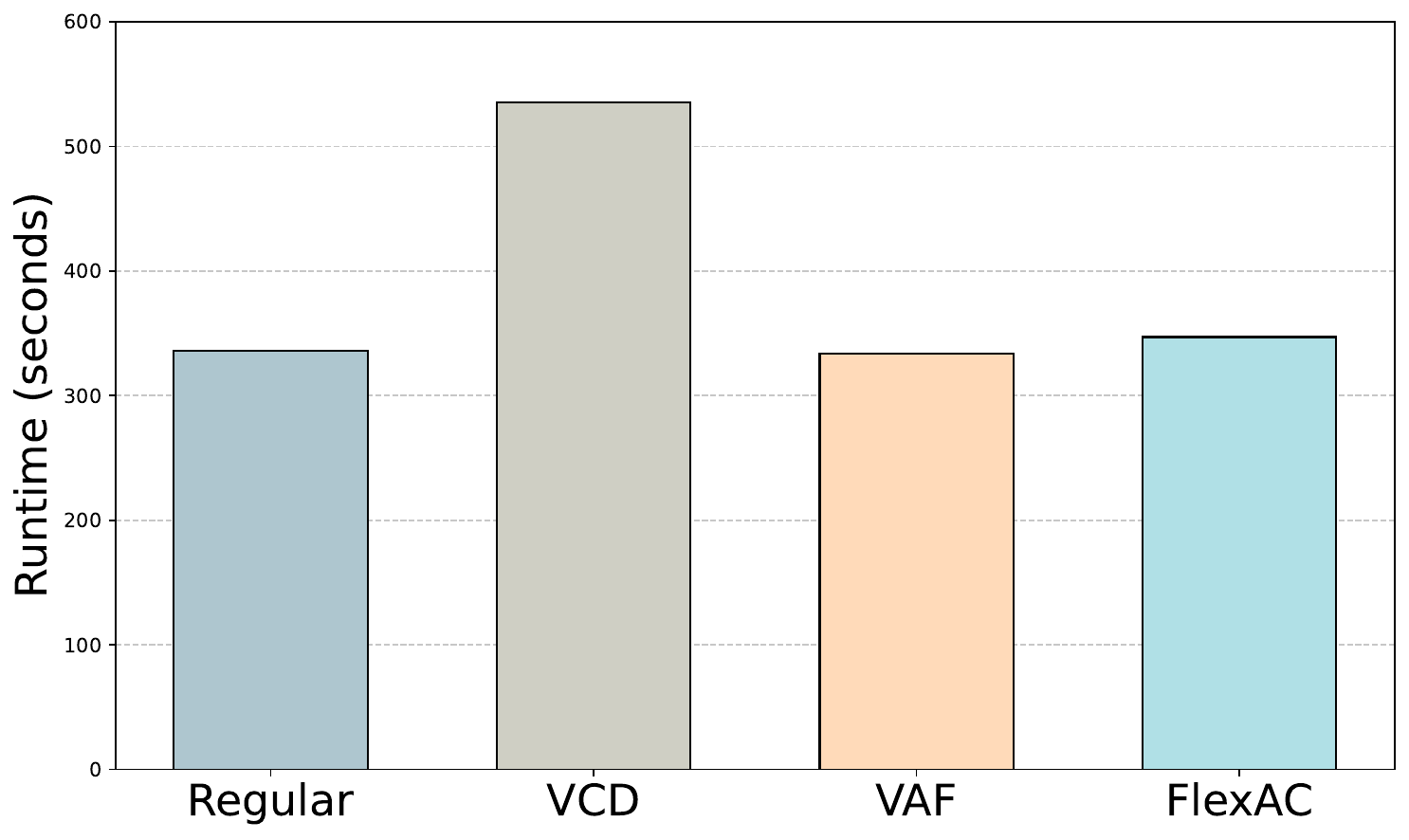}
    \caption{Inference runtime (in seconds) of different methods on Qwen-VL when evaluating CHAIR. FlexAC adds minimal overhead, while VCD incurs high cost due to dual-pass processing.}
    \label{fig_appendix:efficiency}
\end{figure}

To assess the computational efficiency of FlexAC, we compare the inference runtime of different methods on the Qwen-VL model when evaluating the CHAIR benchmark. Specifically, we measure the total time required to process the full test set under each method’s configuration. As shown in ~\Cref{fig_appendix:efficiency}, FlexAC incurs only minimal additional overhead compared to the original model, demonstrating that its control mechanism introduces negligible runtime cost. In contrast, VCD exhibits significantly higher latency due to its reliance on dual forward passes—one for the original image and another for a perturbed version—highlighting its inefficiency. These results confirm that FlexAC achieves controllable reasoning with minimal impact on inference speed.

\subsection{Extended Evaluation on General-Purpose Benchmarks}

To rigorously evaluate FlexAC's impact on general capabilities, we extended our analysis to a diverse suite of 11 benchmarks, as detailed in \Cref{tab:extended_general_benchmarks}. Our evaluation spans three representative categories: \textbf{general multimodal reasoning}, comprising MM-Vet\cite{yu2024mmvet}, MMBench~\cite{liu2024mmbench}, SEED-Bench~\cite{li2023seed}, and MMMB~\cite{sun2025mmmb}; \textbf{vision-centric understanding}, which includes RealWorldQA~\cite{xai2024realworldqa}, CVBench~\cite{zhu2025cvbench}, and AI2D~\cite{kembhavi2016ai2d}; and \textbf{OCR/document-based question answering}, covering TextVQA~\cite{singh2019textvqa}, ChartQA~\cite{masry2022chartqa}, DocVQA~\cite{mathew2021docvqa}, and OCRVQA~\cite{mishra2019ocrvqa}. This comprehensive approach verifies that our control mechanism does not introduce performance degradation.

The results are presented in \Cref{tab:extended_general_benchmarks}. Across all three categories, both FlexAC-C (creativity-enhanced) and FlexAC-P (faithfulness-enhanced) maintain performance closely comparable to the baseline model. This provides strong evidence that our targeted control mechanism effectively modulates associative reasoning without degrading the model's fundamental, general-purpose capabilities.

\begin{table}[h!]
\centering
\caption{Performance of FlexAC on an extended suite of 11 general-purpose benchmarks, grouped by capability. The results demonstrate that FlexAC maintains performance comparable to the baseline across general multimodal, vision-centric, and OCR/document tasks, indicating our method does not harm general capabilities.}
\label{tab:extended_general_benchmarks}
\begin{tabular}{llccc}
\toprule
\textbf{Category} & \textbf{Benchmark} & \textbf{Regular} & \textbf{FlexAC-C} & \textbf{FlexAC-P} \\
\midrule
\multirow{4}{*}{General Multimodal} 
& MM-Vet      & 39.81 & 38.17 & 37.33 \\
& MMBench     & 0.581 & 0.598 & 0.576 \\
& SEED-Bench  & 0.638 & 0.625 & 0.640 \\
& MMMB        & 0.703 & 0.678 & 0.699 \\
\midrule
\multirow{3}{*}{Vision-centric} 
& RealWorldQA & 0.486 & 0.490 & 0.495 \\
& CVBench     & 0.549 & 0.524 & 0.560 \\
& AI2D        & 0.612 & 0.614 & 0.616 \\
\midrule
\multirow{4}{*}{OCR \& Document} 
& TextVQA     & 60.66 & 60.78 & 59.81 \\
& ChartQA     & 48.36 & 49.40 & 45.92 \\
& DocVQA      & 57.79 & 56.85 & 57.59 \\
& OCRVQA      & 47.46 & 49.74 & 45.83 \\
\bottomrule
\end{tabular}
\end{table}

\clearpage
\subsection{Detailed results on POPE}
\label{sec_appendix:pope}
\vspace{-1em}
To complement the summary results in Figure 1, we report detailed POPE evaluation metrics across all settings (random, popular, adversarial) and models in ~\Cref{exp_appendix:pope_detail}. These include accuracy, precision, recall, and F1 scores for all baselines and our FlexAC variants.

\begin{table*}[!ht]
    \vspace{-1em}
    \caption{Performance on POPE. FlexAC here denotes the version configured to suppress associative behavior, aiming to improve factual accuracy (faithfulness).
}
    \vspace{-1em}
    \label{exp_appendix:pope_detail}
    \begin{center}
        \resizebox{0.94\linewidth}{!}{
    \begin{tabular}{c ll cccc}

    \toprule
    \multicolumn{1}{c}{\bf Modal}  & \multicolumn{1}{c}{\bf Setting}  & \multicolumn{1}{c}{\bf Method} & \multicolumn{1}{c}{\bf Accuracy$\uparrow$}  & \multicolumn{1}{c}{\bf Precision $\uparrow$}  & \multicolumn{1}{c}{\bf Recall$\uparrow$} & \multicolumn{1}{c}{\bf F1 Score$\uparrow$} \\

    \midrule

    \multirow{16}{*}{Qwen-VL}  
    & \multirow{4}{*}{\textit{Overall}} 
    & Regular & 86.64 & 92.92 & 79.33 & 85.59 \\
    & & VCD & 87.62 & 91.91 & 82.53 & 86.97 \\
    & & VAF & 87.17 & 91.45 & 82.0 & 86.47 \\
    & & \cellcolor{gray!20} FlexAC (Ours) &\cellcolor{gray!20} 87.44 &\cellcolor{gray!20} 89.31 &\cellcolor{gray!20} 85.07 &\cellcolor{gray!20} \textbf{87.14} \\
    \cmidrule(lr){2-7}
    & \multirow{4}{*}{\textit{random}} 
    & Regular & 88.6 & 97.38 & 79.33 & 87.44 \\
    & & VCD & 89.97 & 97.02 & 82.53 & 89.19 \\
    & & VAF & 89.5 & 96.47 & 82.0 & 88.65 \\
    & & \cellcolor{gray!20} FlexAC (Ours) &\cellcolor{gray!20} 90.0 &\cellcolor{gray!20} 94.38 &\cellcolor{gray!20} 85.07 &\cellcolor{gray!20} \textbf{89.48} \\
    \cmidrule(lr){2-7}
    & \multirow{4}{*}{\textit{popular}} 
    & Regular & 87.0 & 93.7 & 79.33 & 85.92 \\
    & & VCD & 87.97 & 92.6 & 82.53 & 87.28 \\
    & & VAF & 87.7 & 92.55 & 82.0 & 86.96 \\
    & & \cellcolor{gray!20} FlexAC (Ours) &\cellcolor{gray!20} 88.47 &\cellcolor{gray!20} 91.27 &\cellcolor{gray!20} 85.07 &\cellcolor{gray!20} \textbf{88.06} \\
    \cmidrule(lr){2-7}
    & \multirow{4}{*}{\textit{adversarial}} 
    & Regular & 84.33 & 88.15 & 79.33 & 83.51 \\
    & & VCD & 84.93 & 86.69 & 82.53 & \textbf{84.56} \\
    & & VAF & 84.3 & 85.95 & 82.0 & 83.93 \\
    & & \cellcolor{gray!20} FlexAC (Ours) &\cellcolor{gray!20} 83.87 &\cellcolor{gray!20} 83.07 &\cellcolor{gray!20} 85.07 &\cellcolor{gray!20} 84.06 \\
    \midrule

    \multirow{20}{*}{LLaVA-1.5}  
    & \multirow{5}{*}{\textit{Overall}} 
    & Regular & 87.18 & 91.47 & 82.0 & 86.48 \\
    & & HA-DPO & 85.29 & 92.57 & 76.73 & 83.91 \\
    & & VCD & 84.91 & 88.09 & 80.73 & 84.25 \\
    & & VAF & 87.07 & 87.93 & 85.93 & 86.92 \\
    & & \cellcolor{gray!20} FlexAC (Ours) &\cellcolor{gray!20} 87.84 &\cellcolor{gray!20} 87.13 &\cellcolor{gray!20} 88.8 &\cellcolor{gray!20} \textbf{87.96} \\
    \cmidrule(lr){2-7}
    & \multirow{5}{*}{\textit{random}} 
    & Regular & 89.3 & 96.02 & 82.0 & 88.46 \\
    & & HA-DPO & 86.97 & 96.48 & 76.73 & 85.48 \\
    & & VCD & 87.5 & 93.37 & 80.73 & 86.59 \\
    & & VAF & 90.07 & 93.68 & 85.93 & 89.64 \\
    & & \cellcolor{gray!20} FlexAC (Ours) &\cellcolor{gray!20} 91.43 &\cellcolor{gray!20} 93.74 &\cellcolor{gray!20} 88.8 &\cellcolor{gray!20} \textbf{91.2} \\
    \cmidrule(lr){2-7}
    & \multirow{5}{*}{\textit{popular}} 
    & Regular & 87.53 & 92.2 & 82.0 & 86.8 \\
    & & HA-DPO & 86.0 & 94.19 & 76.73 & 84.57 \\
    & & VCD & 85.27 & 88.78 & 80.73 & 84.57 \\
    & & VAF & 87.93 & 89.51 & 85.93 & 87.69 \\
    & & \cellcolor{gray!20} FlexAC (Ours) &\cellcolor{gray!20} 88.7 &\cellcolor{gray!20} 88.62 &\cellcolor{gray!20} 88.8 &\cellcolor{gray!20} \textbf{88.71} \\
    \cmidrule(lr){2-7}
    & \multirow{5}{*}{\textit{adversarial}} 
    & Regular & 84.7 & 86.68 & 82.0 & \textbf{84.28} \\
    & & HA-DPO & 82.9 & 87.53 & 76.73 & 81.78 \\
    & & VCD & 81.97 & 82.78 & 80.73 & 81.74 \\
    & & VAF & 83.2 & 81.48 & 85.93 & 83.65 \\
    & & \cellcolor{gray!20} FlexAC (Ours) &\cellcolor{gray!20} 83.4 &\cellcolor{gray!20} 80.14 &\cellcolor{gray!20} 88.8 &\cellcolor{gray!20} 84.25 \\
    \midrule

    \multirow{16}{*}{DeepSeek-VL}  
    & \multirow{4}{*}{\textit{Overall}} 
    & Regular & 88.42 & 88.13 & 88.8 & 88.47 \\
    & & VCD & 87.82 & 87.64 & 88.07 & 87.85 \\
    & & VAF & 88.37 & 87.59 & 89.4 & 88.49 \\
    & & \cellcolor{gray!20} FlexAC (Ours) &\cellcolor{gray!20} 88.52 &\cellcolor{gray!20} 88.36 &\cellcolor{gray!20} 88.73 &\cellcolor{gray!20} \textbf{88.55} \\
    \cmidrule(lr){2-7}
    & \multirow{4}{*}{\textit{random}} 
    & Regular & 92.0 & 94.87 & 88.8 & \textbf{91.74} \\
    & & VCD & 91.03 & 93.62 & 88.07 & 90.76 \\
    & & VAF & 91.87 & 94.04 & 89.4 & 91.66 \\
    & & \cellcolor{gray!20} FlexAC (Ours) &\cellcolor{gray!20} 91.8 &\cellcolor{gray!20} 94.53 &\cellcolor{gray!20} 88.73 &\cellcolor{gray!20} 91.54 \\
    \cmidrule(lr){2-7}
    & \multirow{4}{*}{\textit{popular}} 
    & Regular & 88.13 & 87.63 & 88.8 & 88.21 \\
    & & VCD & 87.27 & 86.68 & 88.07 & 87.37 \\
    & & VAF & 88.13 & 87.19 & 89.4 & 88.28 \\
    & & \cellcolor{gray!20} FlexAC (Ours) &\cellcolor{gray!20} 88.37 &\cellcolor{gray!20} 88.09 &\cellcolor{gray!20} 88.73 &\cellcolor{gray!20} \textbf{88.41} \\
    \cmidrule(lr){2-7}
    & \multirow{4}{*}{\textit{adversarial}} 
    & Regular & 85.13 & 82.73 & 88.8 & 85.66 \\
    & & VCD & 85.17 & 83.24 & 88.07 & 85.58 \\
    & & VAF & 85.1 & 82.32 & 89.4 & 85.71 \\
    & & \cellcolor{gray!20} FlexAC (Ours) &\cellcolor{gray!20} 85.4 &\cellcolor{gray!20} 83.19 &\cellcolor{gray!20} 88.73 &\cellcolor{gray!20} \textbf{85.87} \\    

    \bottomrule
    \end{tabular}
    }
    \end{center}
    \vspace{-1.5em}
\end{table*}




\section{Ablation Study}

\subsection{Effect of dataset Sizes}
\label{sec_appendix:ablation_sample_num}

To analyze the sensitivity of FlexAC to the number of instances used in control vector construction, we vary Top-K over a wide range: $\{1, 5, 10, 20, 50, 100, 200, 500, 1000, 1500, 2000\}$, and evaluate performance on CHAIRs (↓), CHAIRi (↓), and VDAT (↑) using Qwen-VL.

\begin{figure*}[h]
    \centering
    \subfloat[CHAIR]{
        \includegraphics[width=0.6\textwidth]{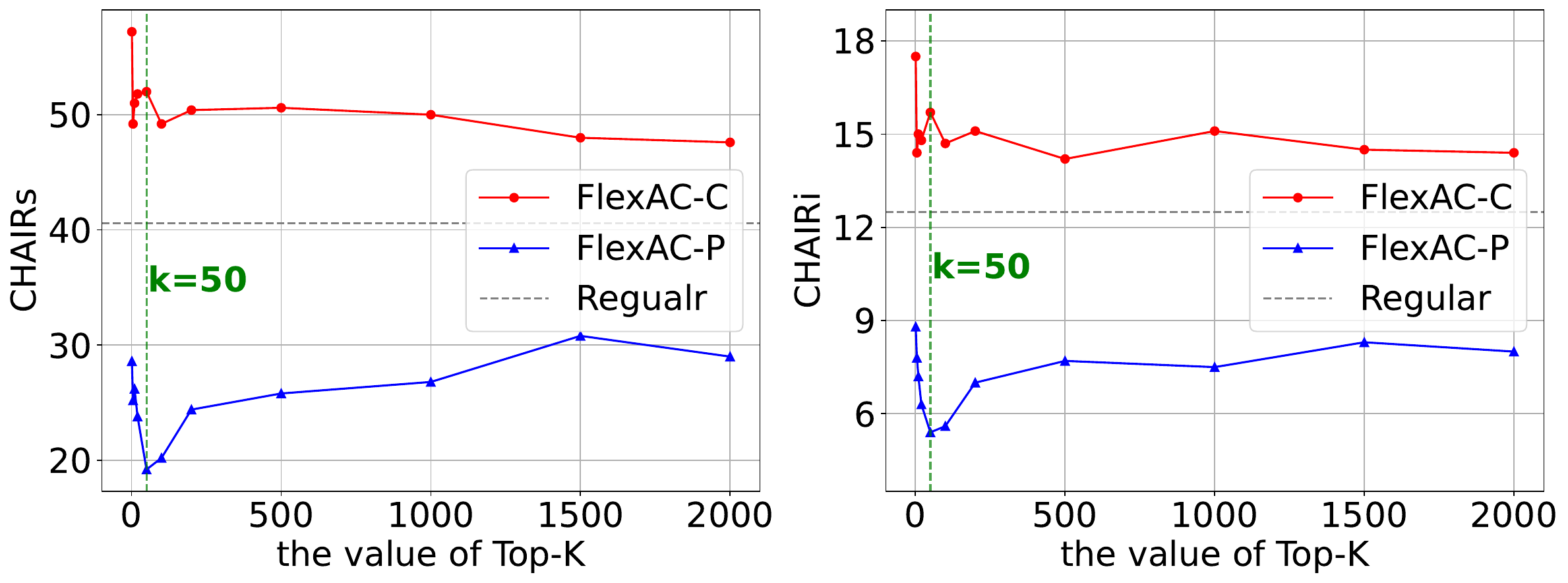}
        \label{fig_appendix:ablation_num_chair}
    }
    \hfill
    \subfloat[VDAT]{
        \includegraphics[width=0.3\textwidth]{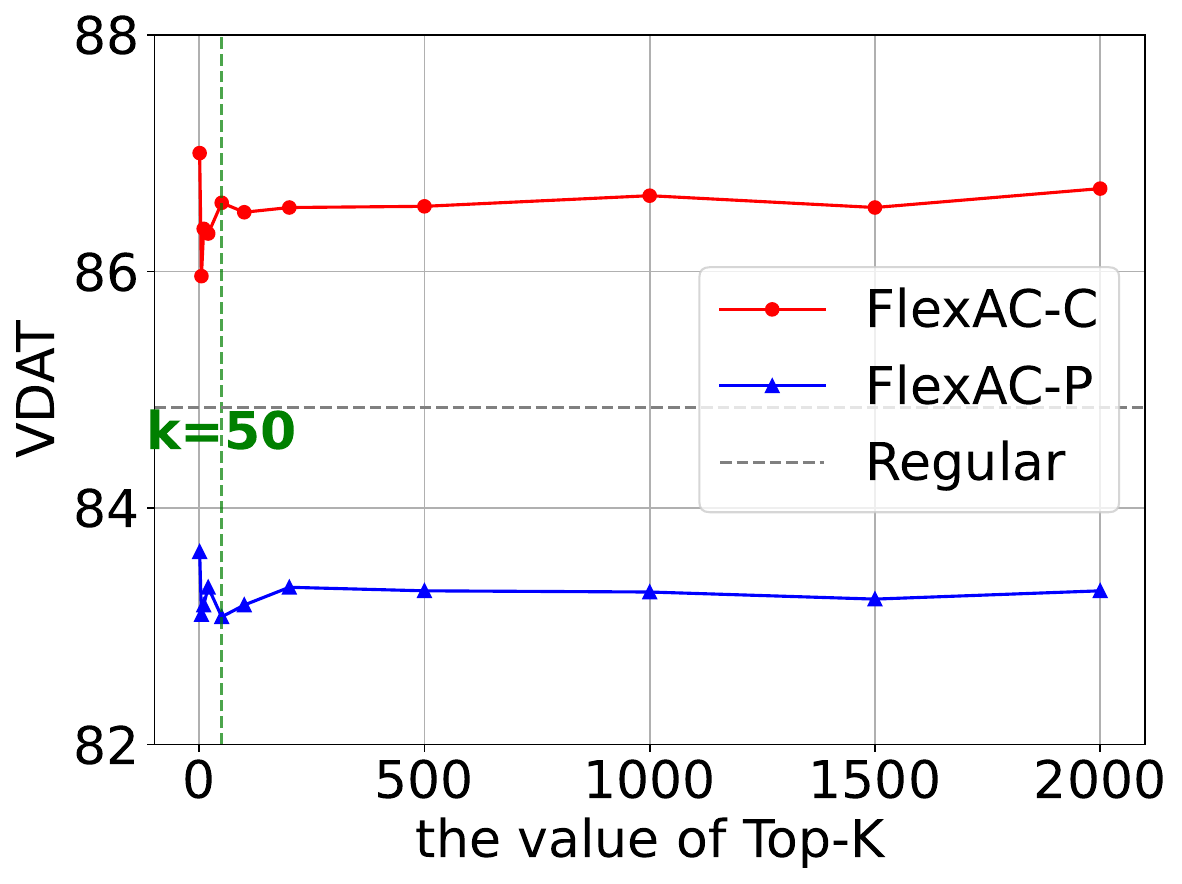}
        \label{fig_appendix:ablation_num_vdat}
    }
    \caption{Sensitivity analysis of the Top-K hyperparameter used in general control vector construction on Qwen-VL. We vary the number of selected instances (K) and evaluate performance on CHAIRs, CHAIRi, and VDAT benchmarks. 
    }
    \label{fig_appendix:ablation_num}

\end{figure*}

As shown in ~\Cref{fig_appendix:ablation_num}, both FlexAC-P (measured on CHAIR for faithfulness) and FlexAC-C (measured on VDAT for creativity) exhibit similar trends: performance is relatively high but unstable when $K$ is very small, and stabilizes near its peak around $K = 50$. Further increasing $K$ leads to slight performance degradation, likely due to noise introduced by instances. 
These results highlight the effectiveness of our Instance Selection strategy, which focuses on selecting a small, high-quality set of associative and non-associative samples.

Notably, across all $K$ values, FlexAC-C and FlexAC-P consistently appear on opposite sides of the Regular baseline, reflecting two associative reasoning strength. This clear separation demonstrates FlexAC’s capacity to bidirectionally modulate reasoning behavior, enabling controllable transitions between creative and faithful outputs.




\clearpage
\subsection{Effect of control layer.}
\label{sec_appendix:ablation_layer}

To validate the generality of our control layer findings beyond Qwen-VL, we conduct additional layer-wise control effectiveness analysis on LLaVA-1.5 and DeepSeek-VL2, as shown in Appendix \Cref{fig_appendix:ablation_layers_llava} and \Cref{fig_appendix:ablation_layers_deepseek}. Similar to the trends observed in Qwen-VL, we find that both FlexAC-C and FlexAC-P exhibit consistent improvements in their respective metrics (VDAT and CHAIR) when applied to middle layers. 
Specifically, the performance peaks around middle layers (layers 10-15) for LLaVA-1.5 and Layers 4-6 for DeepSeek-VL2, which aligns with our feature distance analysis (see Appendix \ref{sec_appendix:feature_distance}). These results further support our choice of control layers and demonstrate that the effectiveness of FlexAC’s modulation strategy generalizes across diverse MLLM architectures.

\begin{figure*}[h]
    \centering
    \subfloat[CHAIR]{
        \includegraphics[width=0.6\textwidth]{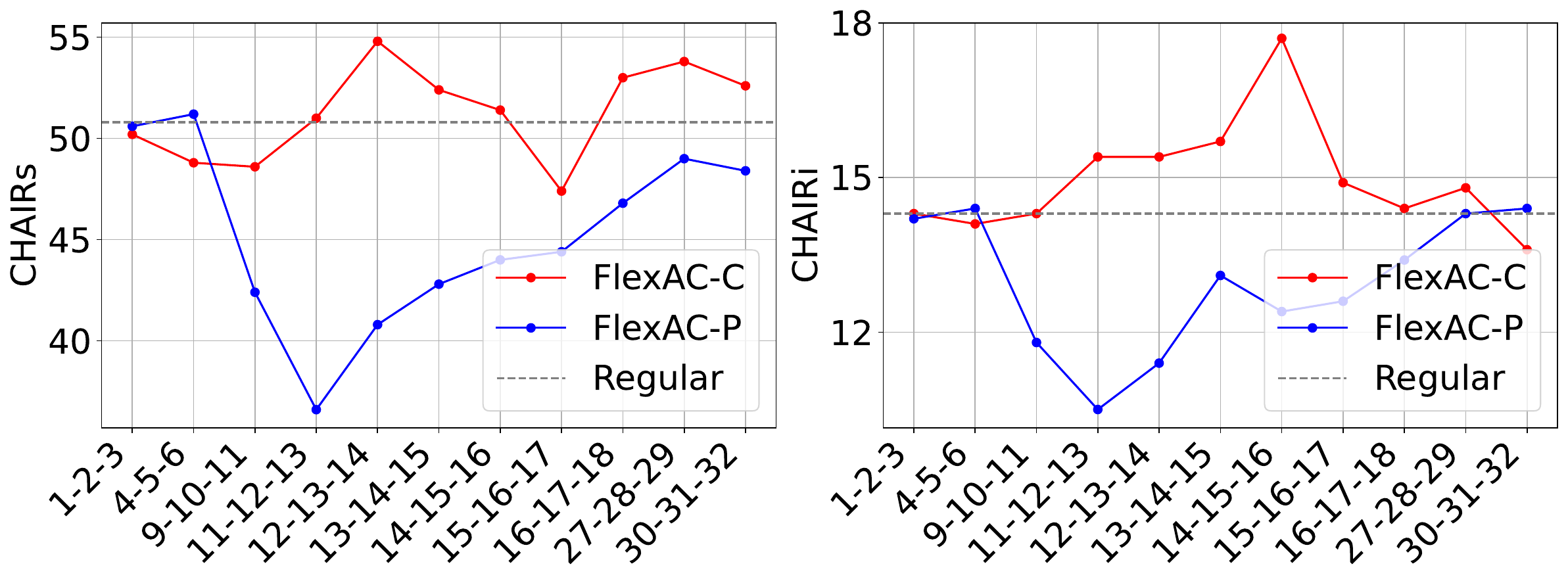}
        \label{fig_appendix:ablation_layers_chair_llava}
    }
    \hfill
    \subfloat[VDAT]{
        \includegraphics[width=0.3\textwidth]{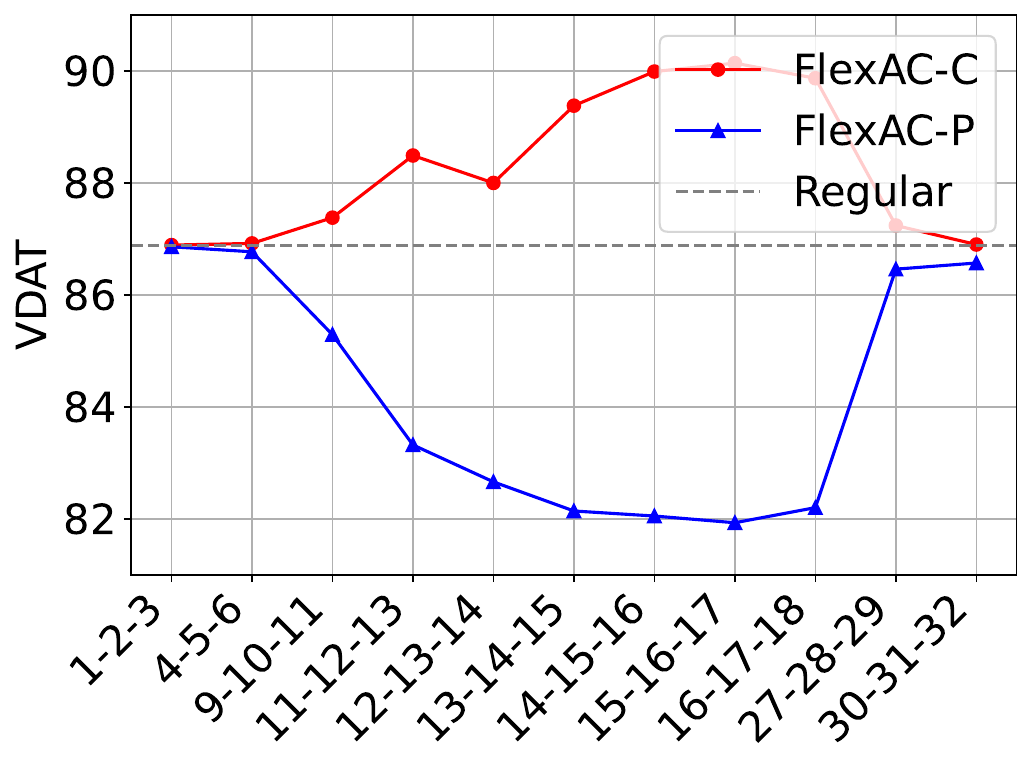}
        \label{fig_appendix:ablation_layers_vdat_llava}
    }
    \caption{Layer-wise analysis of control effectiveness in FlexAC on LLaVA-1.5. The x-axis represents the control layers, while the y-axis shows the performance of the model on CHAIR and VDAT metrics.}
    \label{fig_appendix:ablation_layers_llava}

\end{figure*}

\begin{figure*}[h]
    \centering
    \subfloat[CHAIR]{
        \includegraphics[width=0.6\textwidth]{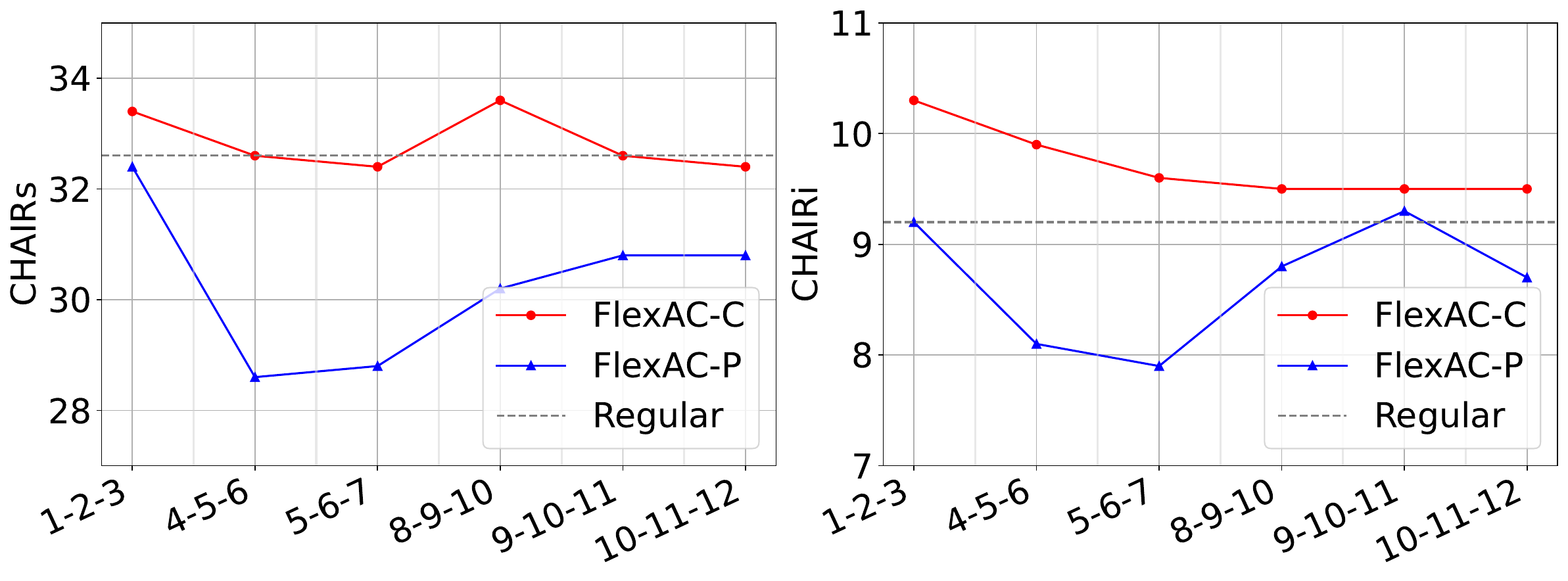}
        \label{fig_appendix:ablation_layers_chair_deepseek}
    }
    \hfill
    \subfloat[VDAT]{
        \includegraphics[width=0.3\textwidth]{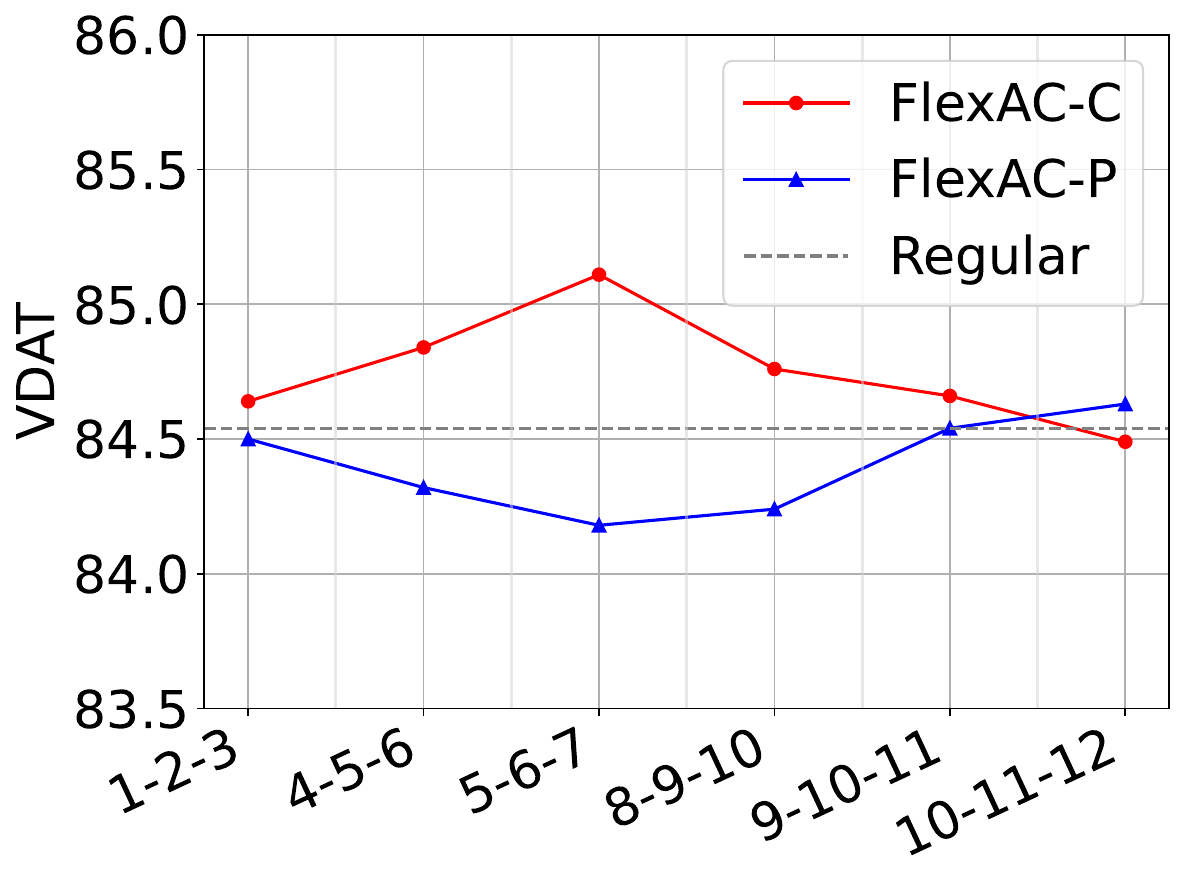}
        \label{fig_appendix:ablation_layers_vdat_deepseek}
    }
    \caption{Layer-wise analysis of control effectiveness in FlexAC on DeepSeek-VL2. The x-axis represents the control layers, while the y-axis shows the performance of the model on CHAIR and VDAT metrics.}
    \label{fig_appendix:ablation_layers_deepseek}

\end{figure*}

\clearpage
\section{Visualizations}

\subsection{Detailed Feature Representation Analysis Using PCA}
\label{sec_appendix:vis_fea_diff}

\begin{figure*}[ht]
    \centering
    \subfloat[]{
        \includegraphics[width=0.98\textwidth]{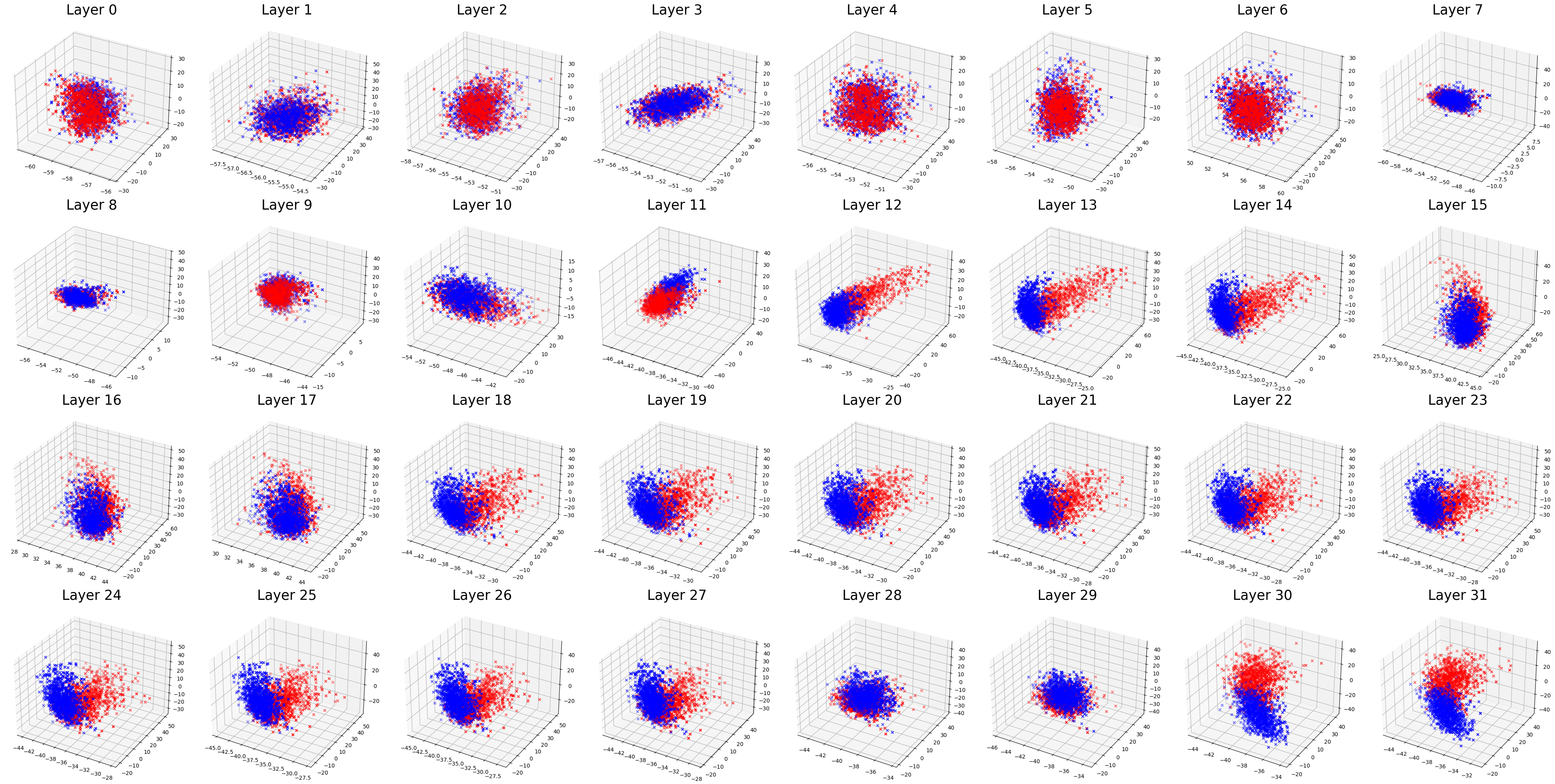}
    }
    \hfill
    \subfloat[]{
        \includegraphics[width=0.98\textwidth]{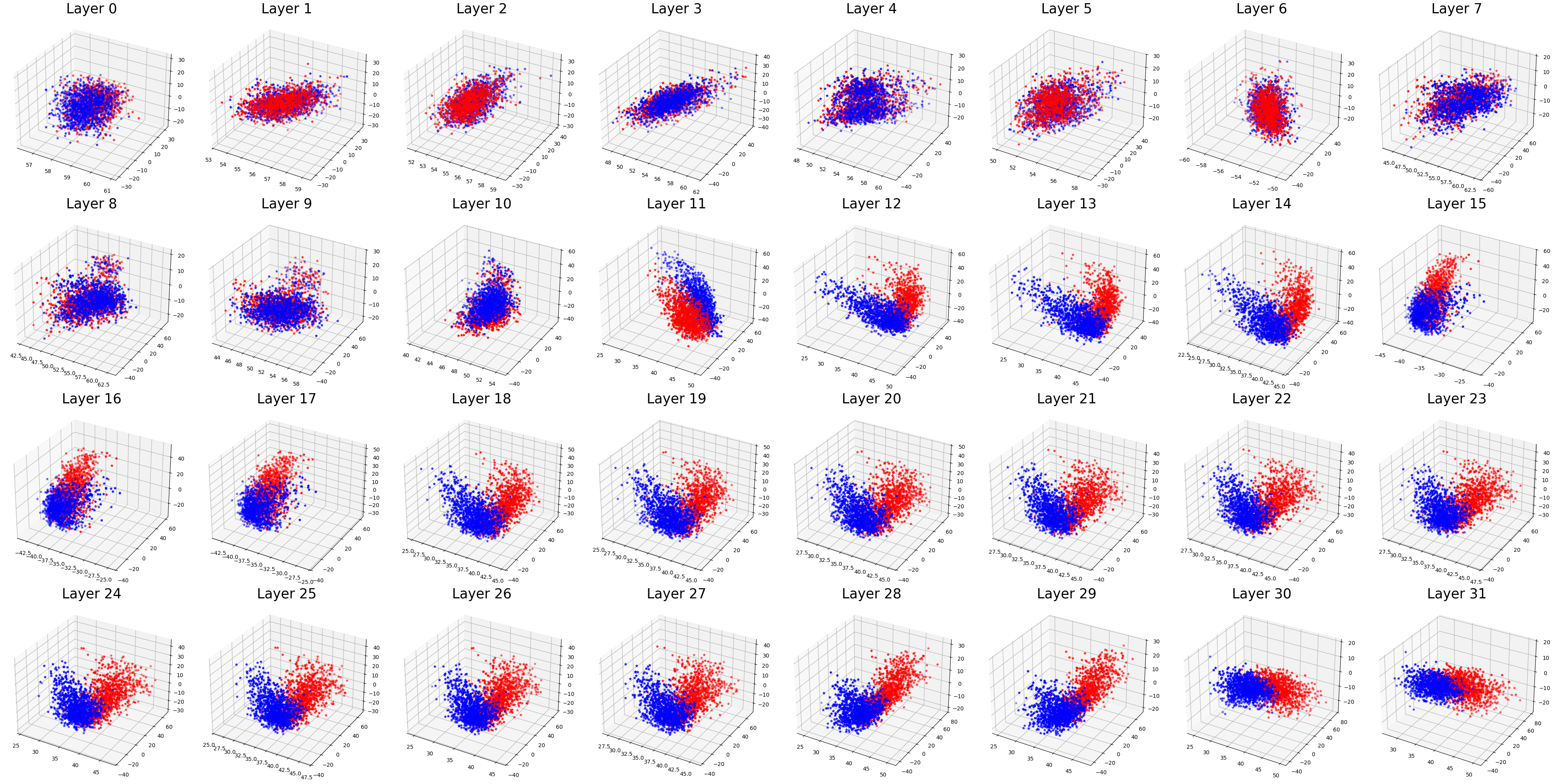}
    }
    \caption{Visualization of feature representations in LLaVA, reduced via PCA, shows the distribution of associative and non-associative data points, represented by red and blue colors, respectively. Subplots (a) and (b) represent the results for different option orders. In deeper layers, the red and blue points exhibit clearer separation, indicating enhanced differentiation between associative and non-associative representations.}
    \label{fig_appendix:diff_ori_fea_all_3d}
\end{figure*}

To provide a more detailed view of how associative and non-associative representations evolve across the model, we present an expanded version of Figure~\textcolor{red}{4} in \Cref{fig_appendix:diff_ori_fea_all_3d}. This visualization shows the PCA-reduced features layer by layer in LLaVA, with red points representing associative features and blue points representing non-associative ones. Compared to the summary visualization, this version reveals how feature separation progressively emerges across layers. In shallow layers (e.g., Layer 0), the two feature types show significant overlap, indicating similar low-level representations. However, starting from the middle layers (around Layer 12), the separation becomes increasingly distinct, highlighting that the model's associative behavior is primarily shaped in the deeper stages.


    

\subsection{Detailed Layer Intervention for Association Localization}
\label{sec_appendix:vis_layer_intervention}

\begin{figure*}[ht]
    \begin{center}
    \includegraphics[width=\textwidth]{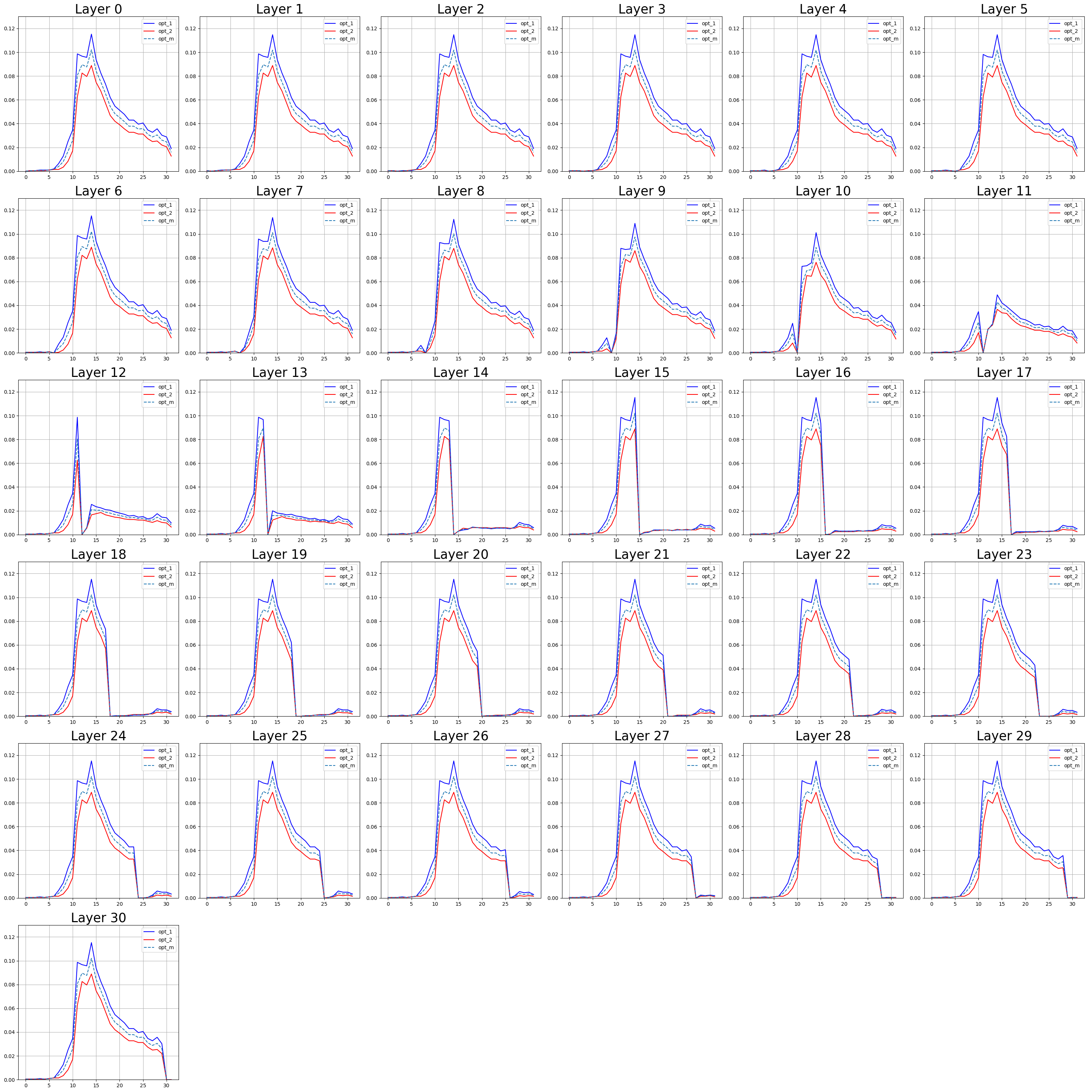}
    \end{center}
    \caption{Feature distance analysis across layers after layer intervention. For example, the subplot titled “Layer 12” shows the feature distances across all layers after replacing associative features at Layer 12 with non-associative features. The X-axis represents the different layers, and the Y-axis represents the \textbf{Cosine distance} between associative and non-associative data.}
    \label{fig:intervention_all_layers_cos}
\end{figure*}

\begin{figure*}[ht]
    \begin{center}
    \includegraphics[width=\textwidth]{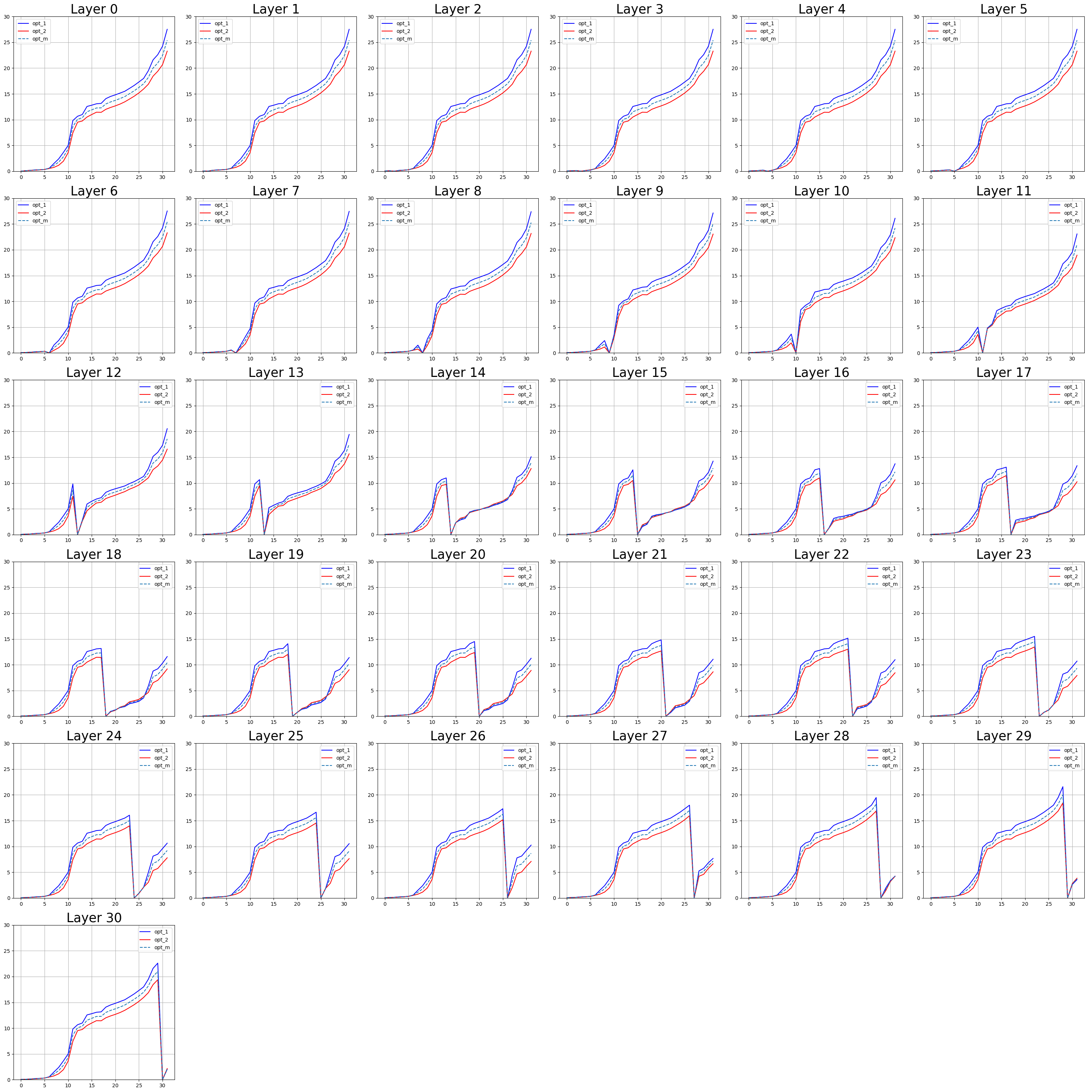}
    \end{center}
    \caption{Feature distance analysis across layers after layer intervention. For example, the subplot titled “Layer 12” shows the feature distances across all layers after replacing associative features at Layer 12 with non-associative features. The X-axis represents the different layers, and the Y-axis represents the \textbf{Euclidean distance} between associative and non-associative data.}
    \label{fig:intervention_all_layers_eu}
\end{figure*}

To gain a more comprehensive understanding of how different layers contribute to associative content generation, we expanded on the analysis presented in Figures \textcolor{red}{2c} and \textcolor{red}{2d} by examining each layer individually. As shown in \Cref{fig:intervention_all_layers_cos} and \Cref{fig:intervention_all_layers_eu}, the detailed version presented here provides a layer-by-layer breakdown of how the interventions affect the model’s internal representations.

In each subplot of this detailed version, we intervened at a specific layer (denoted by the subplot title, e.g., “Layer 0,” “Layer 1,” etc.) by replacing its associative features with non-associative features. We then analyzed the impact of this intervention on feature distances across all layers. 
\Cref{fig:intervention_all_layers_cos} and \Cref{fig:intervention_all_layers_eu} shows that when shallow layers (e.g., layers before Layer 11) are replaced, the feature distances in subsequent layers do not change significantly. However, when middle layers such as Layer 11 are replaced, the subsequent feature distances drop sharply, indicating that these layers have a crucial impact on the model’s associative tendencies. In contrast, when deeper layers (e.g., layers after Layer 14) are replaced, the changes in subsequent layers become more stable, suggesting that deeper layers have a weaker influence on associative tendencies.

This detailed analysis highlights that replacing features at specific layers has a distinct influence on subsequent layers, with the greatest impact often observed in middle layers. This is consistent with the averaged results in Figures \textcolor{red}{2c} and \textcolor{red}{2d}, which pointed towards the critical role of middle layers in maintaining associative characteristics.

\clearpage
\subsection{Visualization of more examples}
\label{sec_appendix:more_vis_examples}

\begin{figure}[h]
    \centering
    \includegraphics[width=0.99\linewidth]{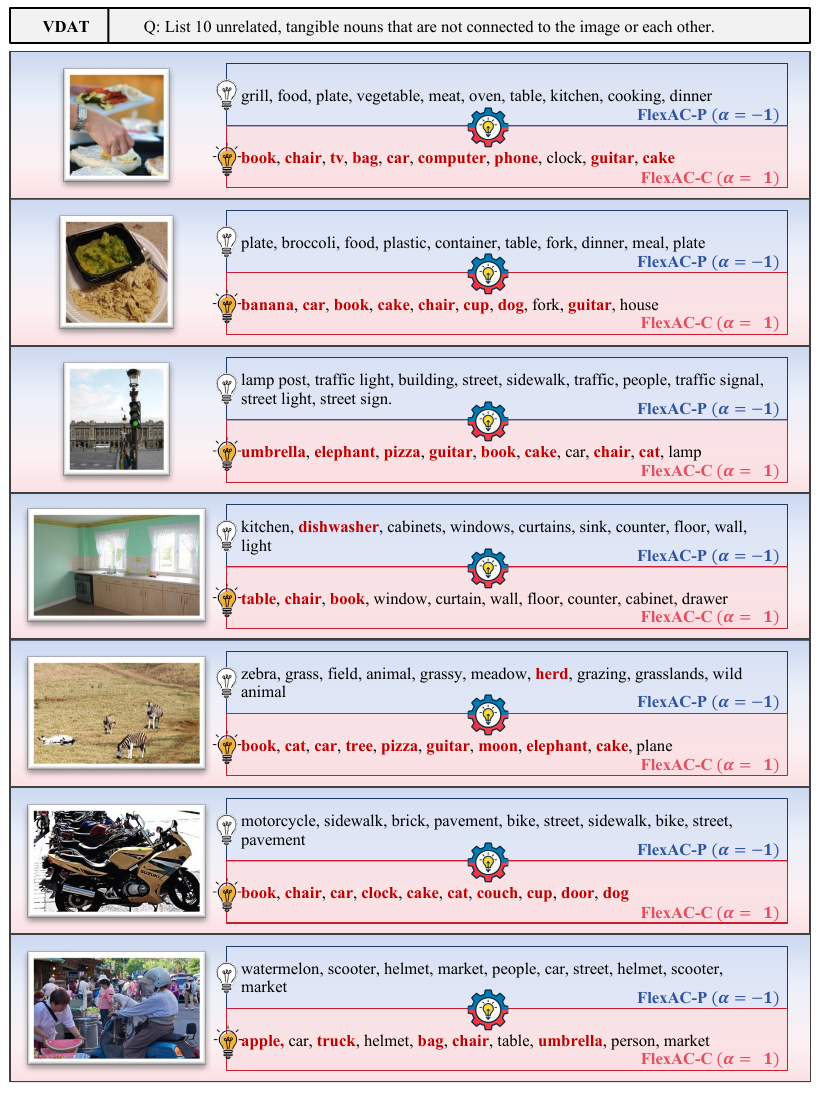}
    \caption{\textbf{Visualization of FlexAC’s Control on VDAT}, showing how FlexAC-P (faithfulness) and FlexAC-C (creativity) adjust the level of associative strength in the generated responses.
    }
    \label{fig_appendix:vdat_demo}
\end{figure}

\begin{figure}[h]
    \centering
    \includegraphics[width=0.99\linewidth]{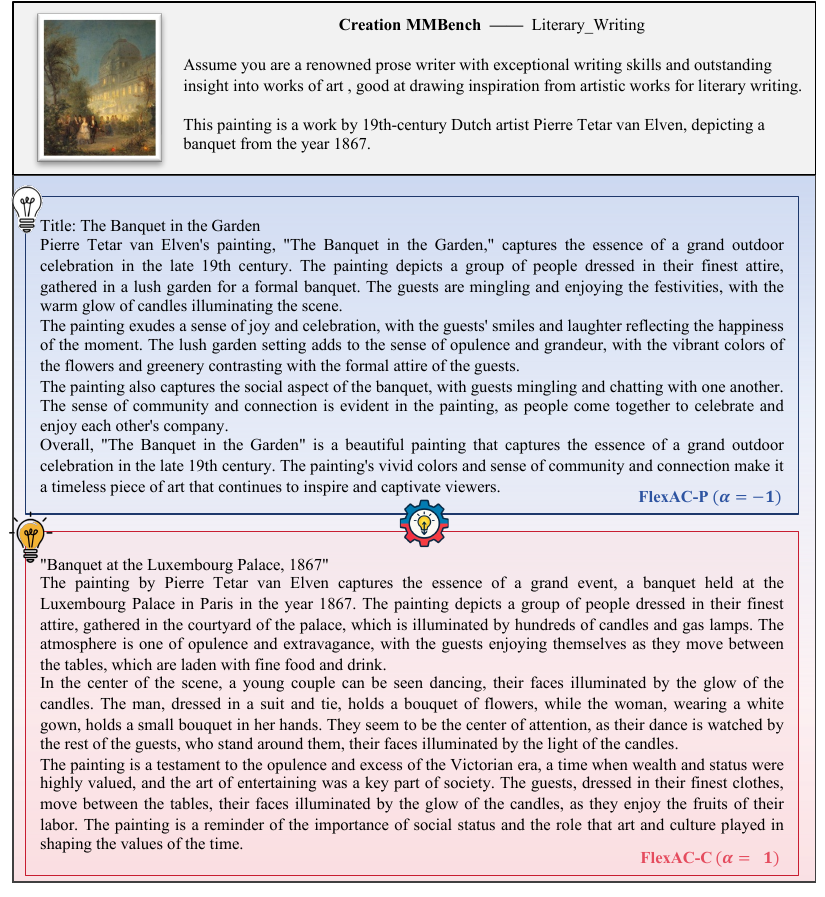}
    \caption{\textbf{Visualization of FlexAC’s Control on Creation MMBench}, showing how FlexAC-P (faithfulness) and FlexAC-C (creativity) adjust the level of associative strength in the generated responses.
    }
    \label{fig_appendix:creation_demo1}
\end{figure}

\begin{figure}[h]
    \centering
    \includegraphics[width=0.99\linewidth]{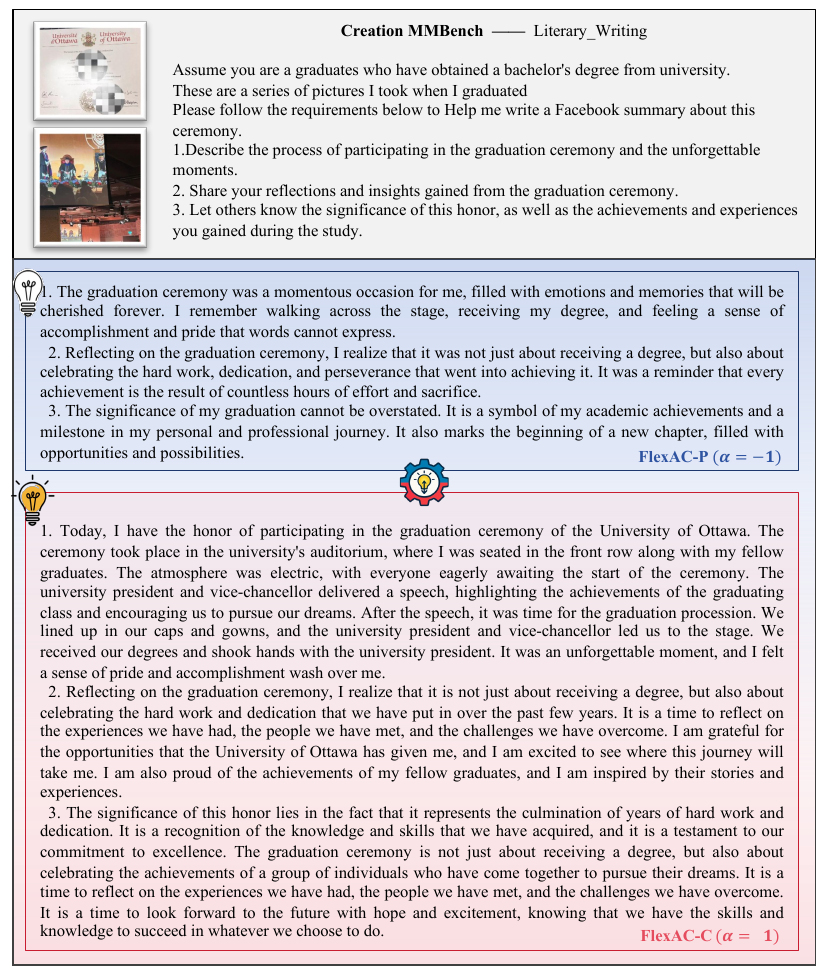}
    \caption{\textbf{Visualization of FlexAC’s Control on Creation MMBench}, showing how FlexAC-P (faithfulness) and FlexAC-C (creativity) adjust the level of associative strength in the generated responses.
    }
    \label{fig_appendix:creation_demo2}
\end{figure}

\begin{figure}[h]
    \centering
    \includegraphics[width=0.99\linewidth]{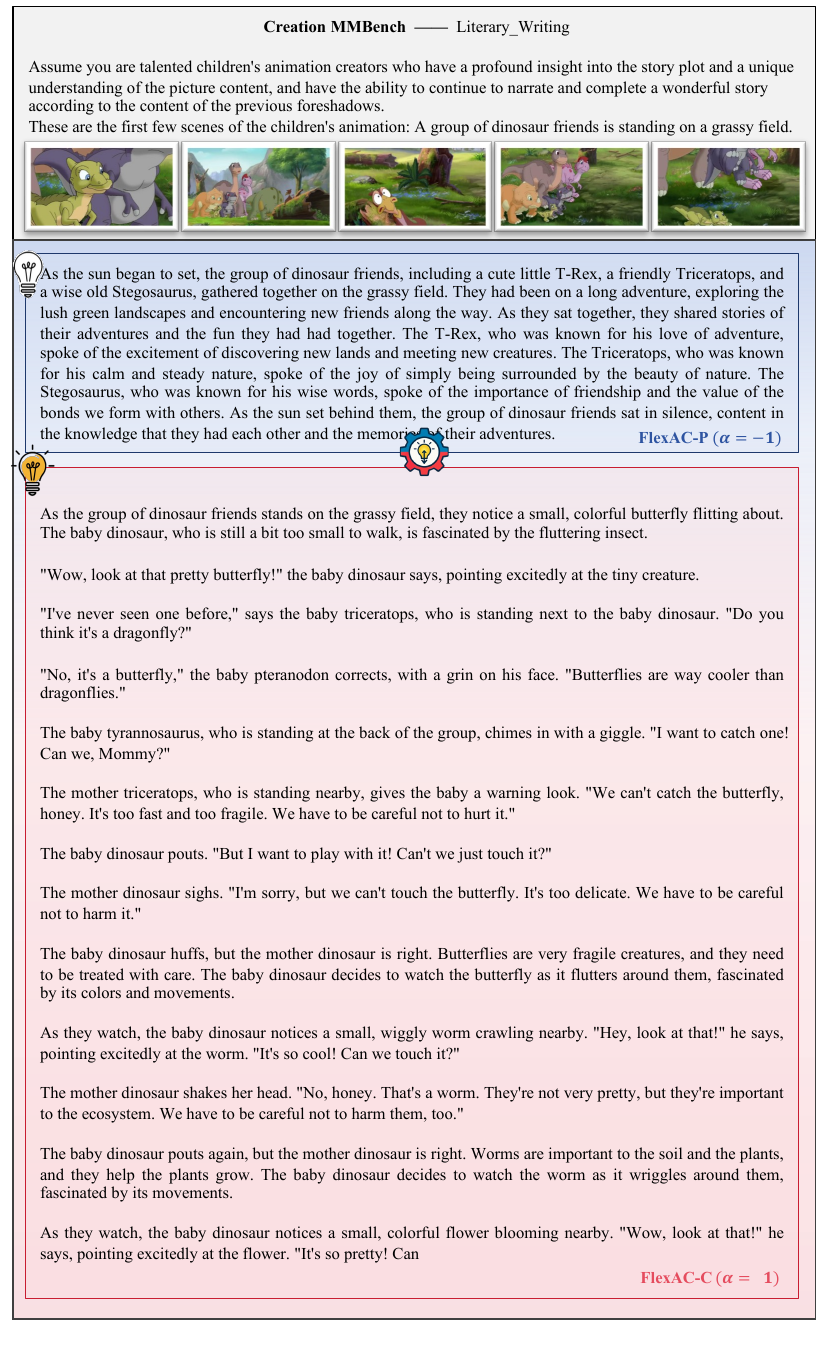}
    \caption{\textbf{Visualization of FlexAC’s Control on Creation MMBench}, showing how FlexAC-P (faithfulness) and FlexAC-C (creativity) adjust the level of associative strength in the generated responses.
    }
    \label{fig_appendix:creation_demo3}
\end{figure}

\begin{figure}[h]
    \centering
    \includegraphics[width=0.99\linewidth]{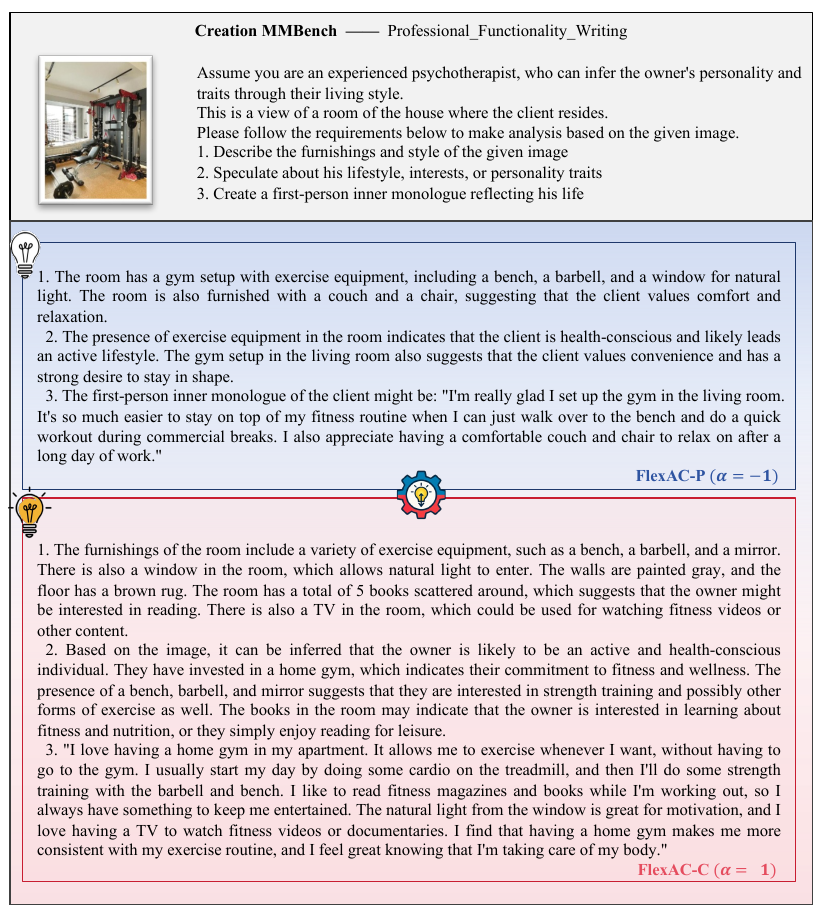}
    \caption{\textbf{Visualization of FlexAC’s Control on Creation MMBench}, showing how FlexAC-P (faithfulness) and FlexAC-C (creativity) adjust the level of associative strength in the generated responses.
    }
    \label{fig_appendix:creation_demo4}
\end{figure}

\begin{figure}[h]
    \centering
    \includegraphics[width=0.99\linewidth]{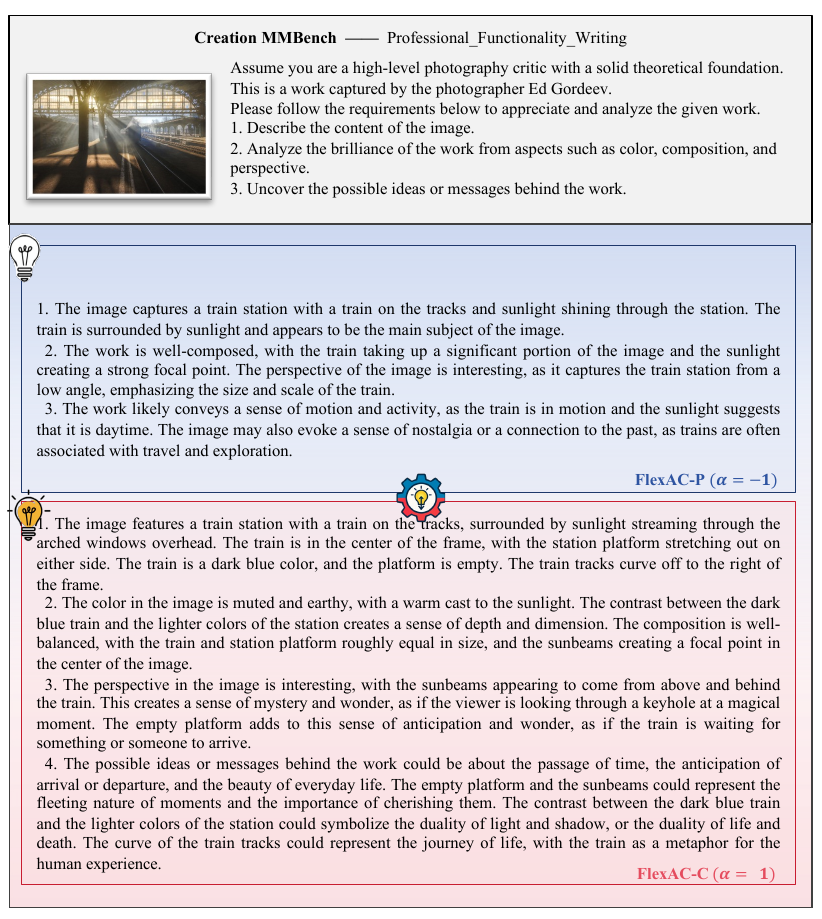}
    \caption{\textbf{Visualization of FlexAC’s Control on Creation MMBench}, showing how FlexAC-P (faithfulness) and FlexAC-C (creativity) adjust the level of associative strength in the generated responses.
    }
    \label{fig_appendix:creation_demo5}
\end{figure}

\begin{figure}[h]
    \centering
    \includegraphics[width=0.99\linewidth, height=0.85\textheight, keepaspectratio]{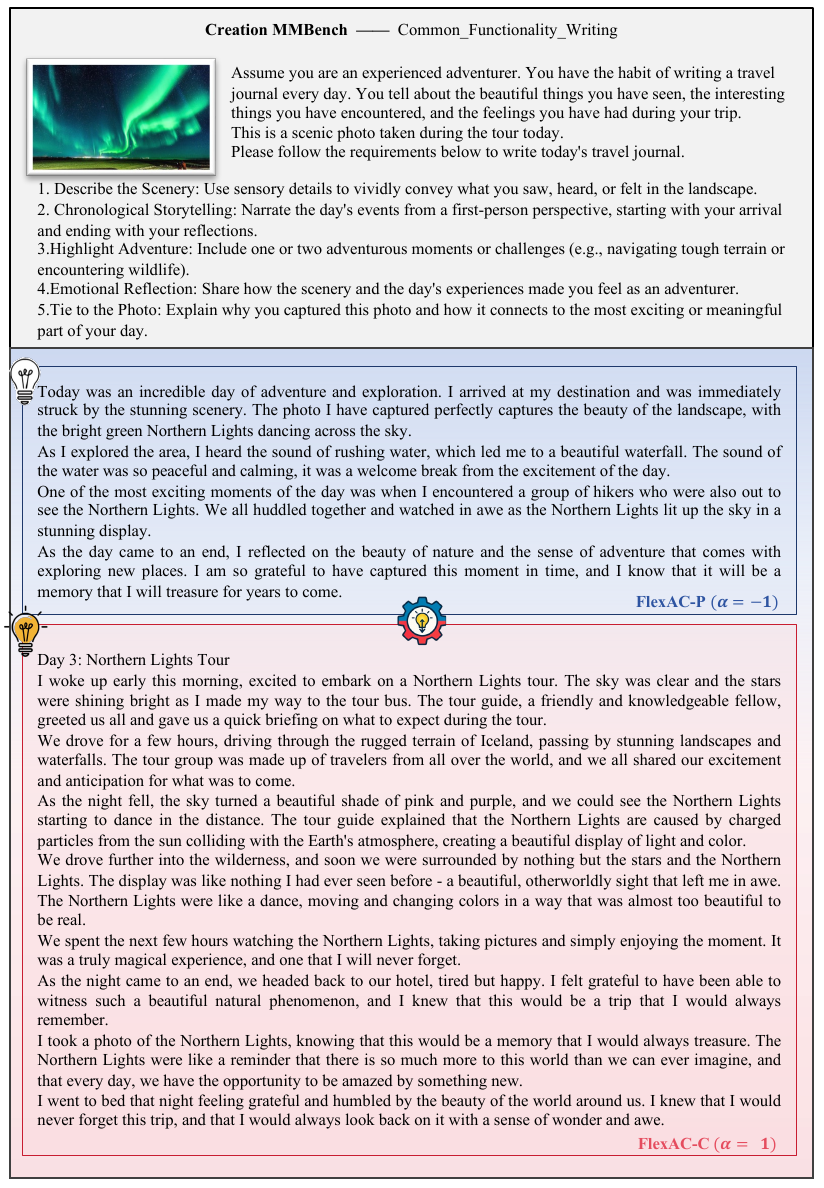}
    \caption{\textbf{Visualization of FlexAC’s Control on Creation MMBench}, showing how FlexAC-P (faithfulness) and FlexAC-C (creativity) adjust the level of associative strength in the generated responses.
    }
    \label{fig_appendix:creation_demo6}
\end{figure}

We visualize sample outputs from both the creativity- and precision-enhancing variants of FlexAC on Qwen-VL. \Cref{fig_appendix:vdat_demo} shows examples from the VDAT benchmark, highlighting differences in associative strength. Additional qualitative results on Creation-MMBench are provided in Figures~\ref{fig_appendix:creation_demo1}–\ref{fig_appendix:creation_demo6}, illustrating FlexAC’s ability to adjust associative reasoning across creative tasks.



\clearpage
\subsection{Feature Distance Analysis on Additional Models}
\label{sec_appendix:feature_distance}
\begin{figure}[h]
    \centering
    \subfloat[Qwen-VL]{
        \includegraphics[width=0.45\linewidth]{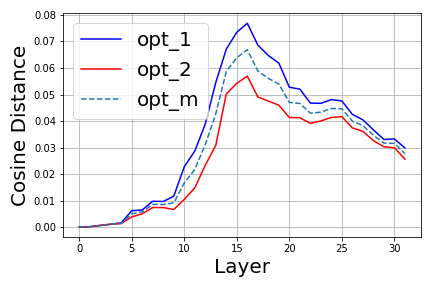}
        \label{fig_appendix:feature_diff_1}
    }
    \hfill
    \subfloat[Qwen-VL]{
        \includegraphics[width=0.45\linewidth]{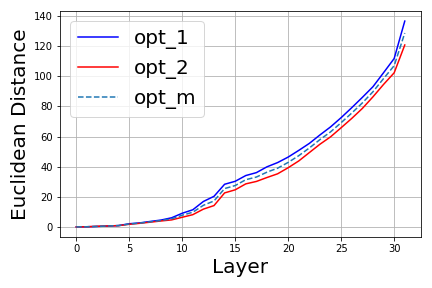}
        \label{fig_appendix:feature_diff_2}
    }
    \\
    \subfloat[Deepseek-VL2]{
        \includegraphics[width=0.45\linewidth]{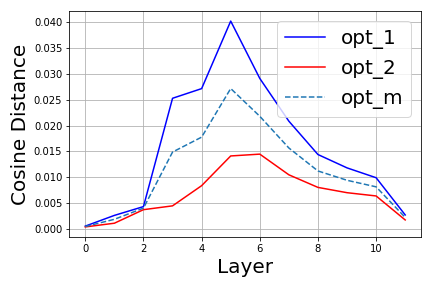}
        \label{fig_appendix:feature_diff_3}
    }
    \hfill
    \subfloat[Deepseek-VL2]{
        \includegraphics[width=0.45\linewidth]{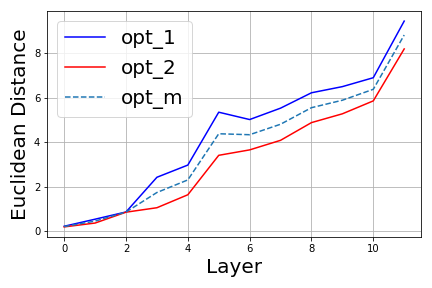}
        \label{fig_appendix:feature_diff_4}
    }
    \caption{Layer-wise feature distance trends between associative and non-associative representations on Qwen-VL and Deepseek-VL2, extending the LLaVA results from Section 3.1.
    }
    \label{fig_appendix:feature_diff}

\end{figure}

To complement the analysis in Section 3.1, we extend the feature distance evaluation to two additional MLLMs: Qwen-VL and Deepseek-VL2. As in the main study, we compute the cosine and Euclidean distances between associative and non-associative representations extracted from each transformer layer. The results are shown in Figure~\ref{fig_appendix:feature_diff}.

Consistent with our findings on LLaVA, we observe that cosine distance peaks in the middle layers, while Euclidean distance gradually increases throughout the network. These patterns reinforce the conclusion that associative behavior primarily emerges and diverges in the middle layers, while deep layers largely propagate those effects.


Importantly, this analysis also informs the design of our control strategy. In Qwen-VL, the middle layers are approximately $13-20$, and in DeepSeek-VL2, $3-7$. Accordingly, we select Layers $15-17$ for Qwen-VL and Layers $4-6$ for DeepSeek-VL2 as control points in FlexAC. These ranges correspond to the regions of maximal divergence between associative and non-associative features, enabling targeted yet lightweight intervention.

